\documentclass[10pt,twocolumn,letterpaper]{article}

\usepackage{cvm}
\usepackage{times}
\usepackage{epsfig}
\usepackage{graphicx}
\usepackage{amsmath}
\usepackage{amssymb}
\usepackage{paralist,bbding,pifont}
\usepackage{subfigure}

\usepackage{multirow}
\usepackage{bbm}
\usepackage{booktabs}

\usepackage[pagebackref=true,breaklinks=true,colorlinks,bookmarks=false]{hyperref}
\usepackage{algorithm, algorithmic}

\newcommand{\keywords}[1]{{\bf \emph{Keywords: #1}}}

\newcommand{\tablestyle}[2]{\setlength{\tabcolsep}{#1}\renewcommand{\arraystretch}{#2}\centering\footnotesize}

\newlength\savewidth\newcommand\shline{\noalign{\global\savewidth\arrayrulewidth
		\global\arrayrulewidth 1pt}\hline\noalign{\global\arrayrulewidth\savewidth}}

\cvmfinalcopy %

\def\cvmPaperID{210} %

\ifcvmfinal\fi
\begin{document}

\title{Semantic Segmentation via Pixel-to-Center Similarity Calculation}

\author{
Dongyue Wu\textsuperscript{1}, 
Zilin Guo\textsuperscript{1}, Aoyan Li\textsuperscript{1}, Changqian Yu\textsuperscript{2}, Changxin Gao\textsuperscript{1}, Nong Sang\textsuperscript{1}
\\
\textsuperscript{1}National Key Laboratory of Science and Technology on Multispectral Information Processing, School of \\ Artificial
Intelligence and Automation, Huazhong University of Science and Technology, Wuhan, China\\
{\tt\small \{dongyue\_wu,zilin\_guo,aoyanli,cgao,nsang\}@hust.edu.cn } \\
\textsuperscript{2}Meituan Inc., Beijing, China\\
{\tt\small changqianyu@meituan.com} 

}

\maketitle

\begin{abstract}

Since the fully convolutional network has achieved great success in semantic segmentation, lots of works have been proposed focusing on extracting discriminative pixel feature representations. 
However, we observe that existing methods still suffer from two typical challenges, 
\ie (i) large intra-class feature variation in different scenes, 
(ii) small inter-class feature distinction in the same scene.
In this paper, we first rethink semantic segmentation from a perspective of similarity between pixels and class centers.
Each weight vector of the segmentation head represents its corresponding semantic class in the whole dataset, which can be regarded as the embedding of the class center.
Thus, the pixel-wise classification amounts to computing similarity in the final feature space between pixels and the class centers.
Under this novel view, we propose a Class Center Similarity layer~(CCS layer) to address the above-mentioned challenges by generating adaptive class centers conditioned on different scenes and supervising the similarities between class centers.
It utilizes a Adaptive Class Center Module (ACCM) to generate class centers conditioned on each scene, which adapt the large intra-class variation between different scenes.
Specially designed loss functions are introduced to control both inter-class and intra-class distances based on predicted center-to-center and pixel-to-center similarity, respectively.
Finally, the CCS layer outputs the processed pixel-to-center similarity as the   segmentation prediction.
Extensive experiments demonstrate that our model performs favourably 
against the state-of-the-art CNN-based methods. 

\end{abstract}

\section{Introduction}
\label{sec:intro}

\begin{figure*}[htbp]
\centering
\includegraphics[width=0.95\linewidth]{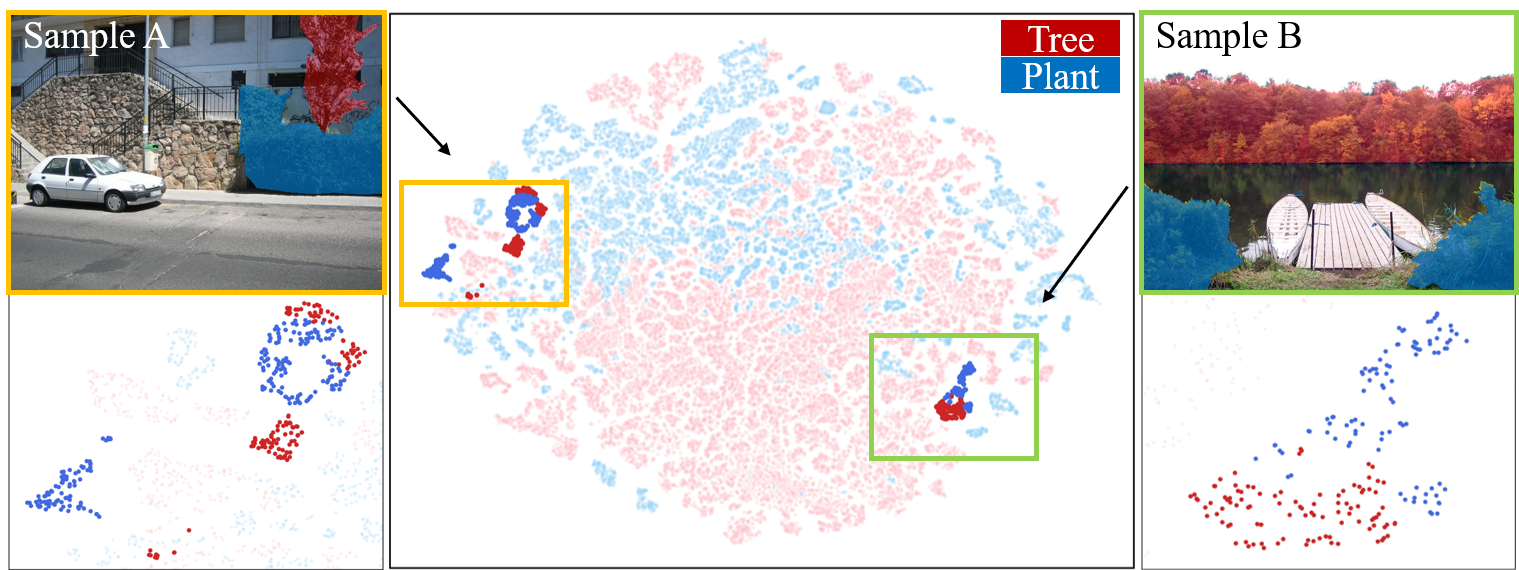}
\caption{
Illustration of the overall distribution of ``tree" and ``plant" pixels on the ADE20K validation set.
We show the distribution of the two classes in feature space.
For clarity, we only label the pixels belonging to ``tree" and ``plant", which are in red and blue, respectively, while the pixels that belong to other classes keep their original color. Each dot in our plot represents a randomly sampled pixel in the feature space. The light-colored dots denote pixels sampled from other samples(scenes) in the whole dataset, while the dark-colored ones are sampled from \textit{Sample A} and \textit{Sample B}.
According to the plots, the features of both ``tree" and ``plant" in \textit{Sample A} are quite different from those in \textit{Sample B}, suggesting large intra-class variation between different scenes. The features of ``tree" and ``plant" are hard to distinguish in \textit{Sample A}, because of the small inter-class distinction. The dimension of features is reduced for illustration using t-SNE~\cite{tsne}.
}
\label{samplefig}
\end{figure*}
Semantic segmentation aims to assign each pixel with a semantic category,
which is a fundamental and challenging task in the computer vision field.
Benefited from the development of deep convolutional networks~\cite{simonyan2014very,krizhevsky2012imagenet,russakovsky2015imagenet,huang2017densely},
the fully convolutional network~(FCN)~\cite{long2015fully}
has been the dominant solution in the semantic segmentation task.

\begin{figure}[htb]
\footnotesize
\centering
\renewcommand{\tabcolsep}{1pt} %
\renewcommand{\arraystretch}{1} %
\begin{center}
\begin{tabular}{ccccc}
\includegraphics[width=0.3\linewidth]{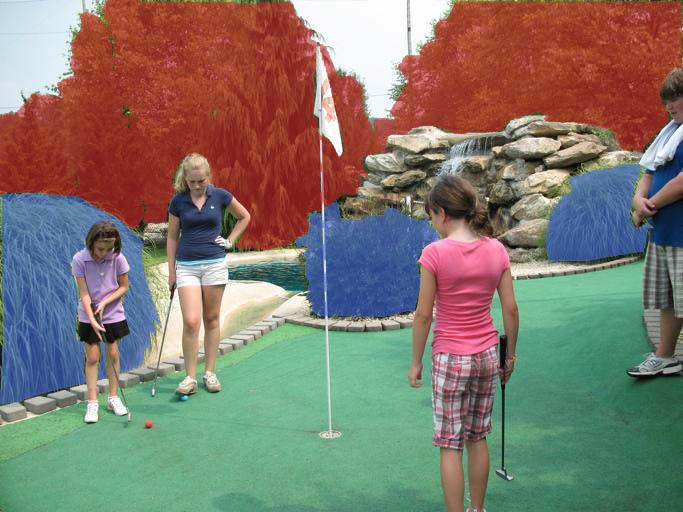} &
\includegraphics[width=0.3\linewidth]{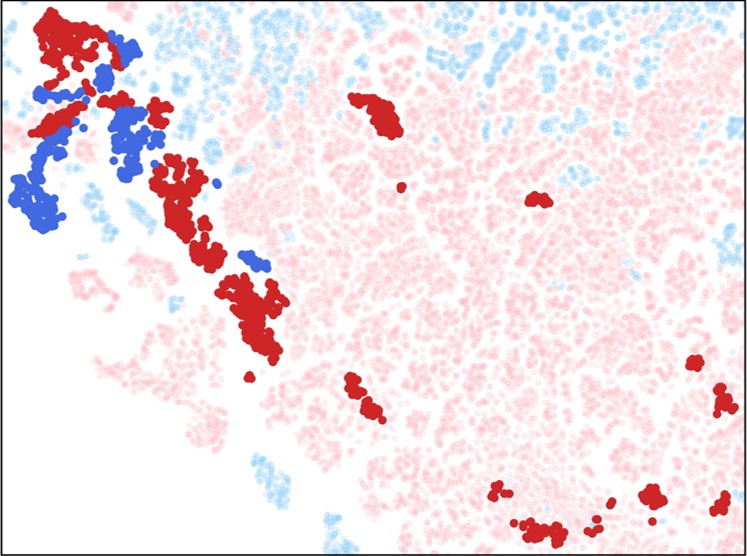} &
\includegraphics[width=0.3\linewidth]{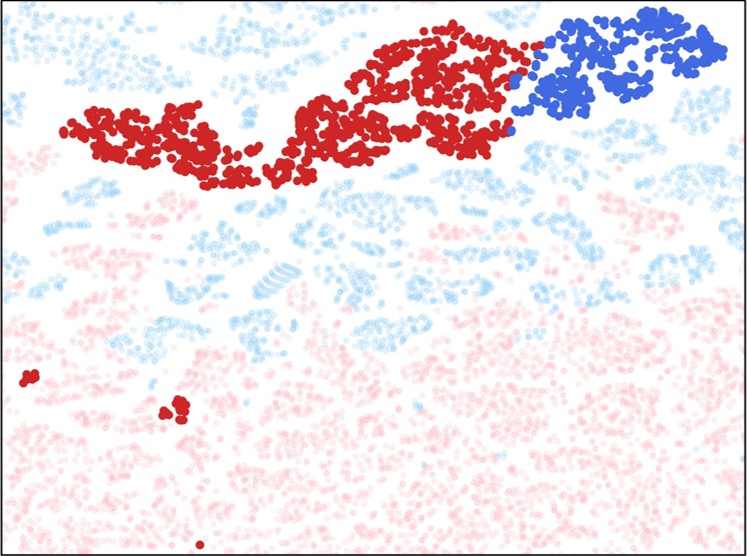}\\
\includegraphics[width=0.3\linewidth]{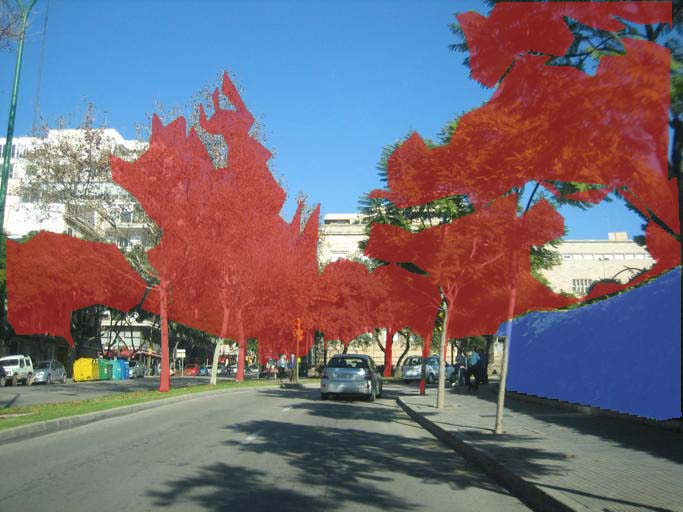} &
\includegraphics[width=0.3\linewidth]{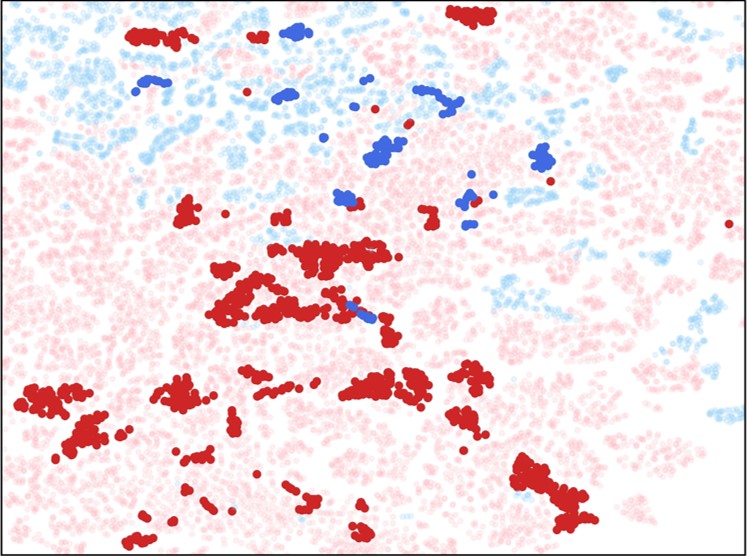} &
\includegraphics[width=0.3\linewidth]{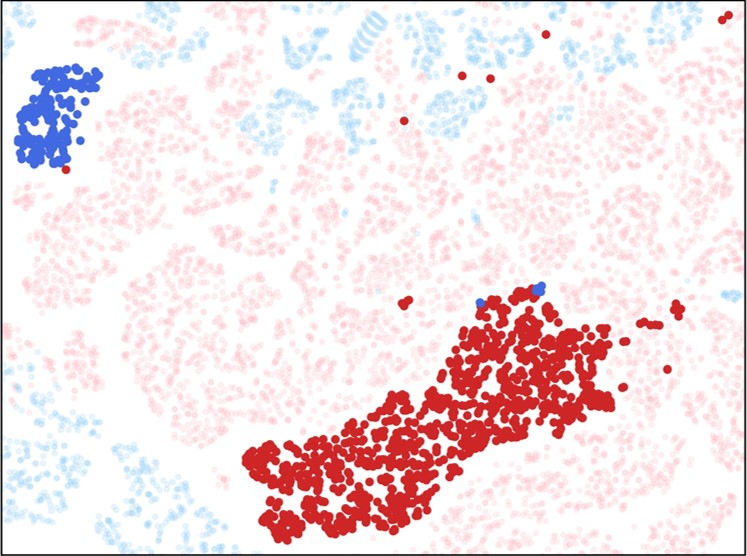}\\
\includegraphics[width=0.3\linewidth]{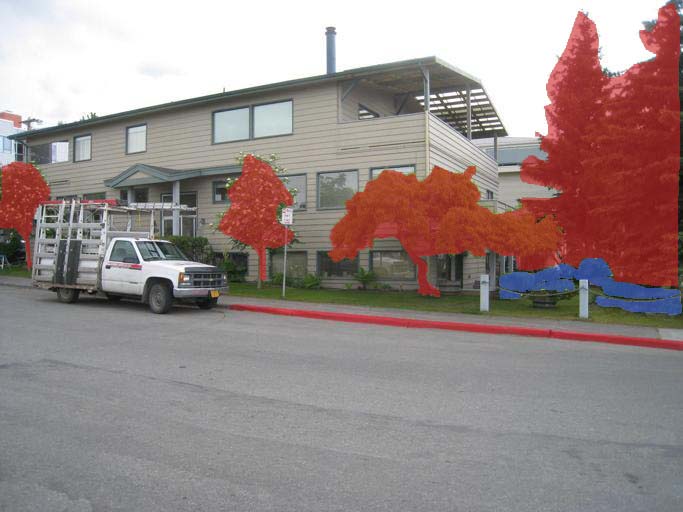} &
\includegraphics[width=0.3\linewidth]{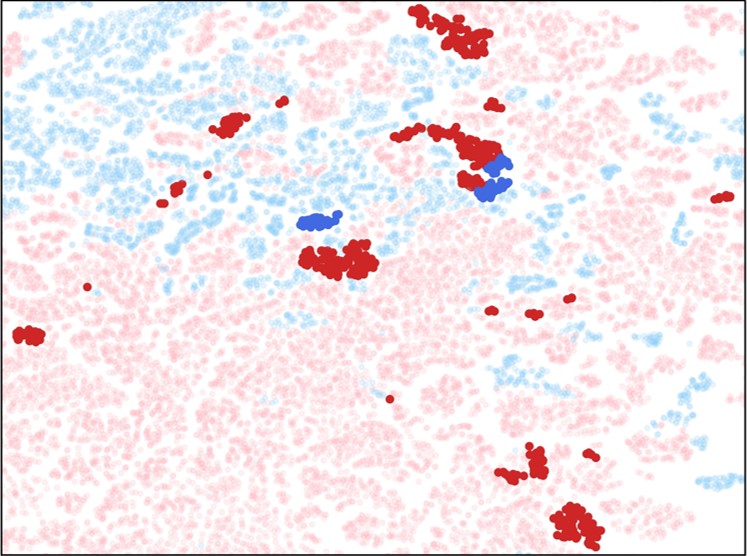} &
\includegraphics[width=0.3\linewidth]{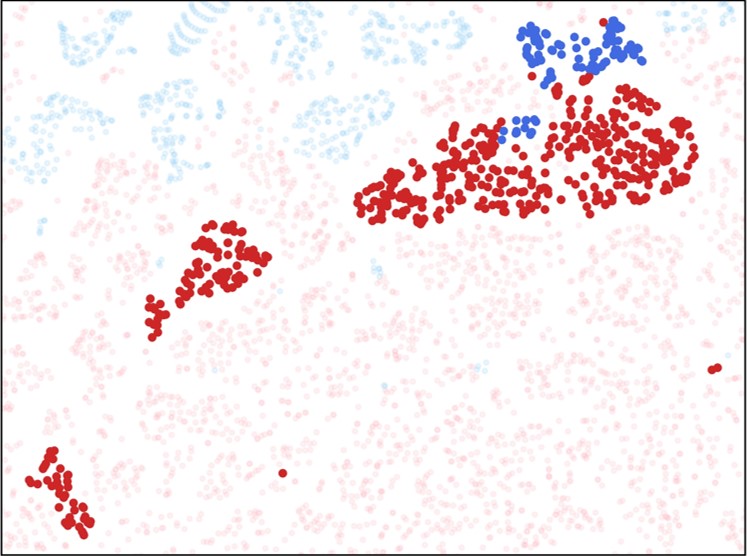}\\
\includegraphics[width=0.3\linewidth]{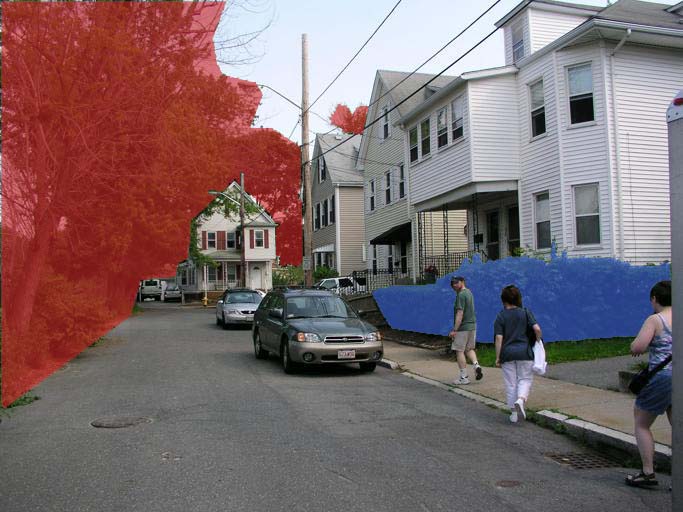} &
\includegraphics[width=0.3\linewidth]{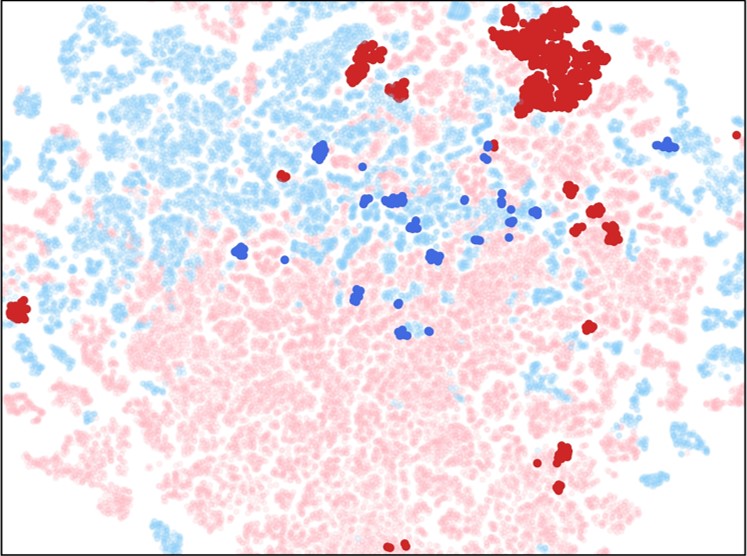} &
\includegraphics[width=0.3\linewidth]{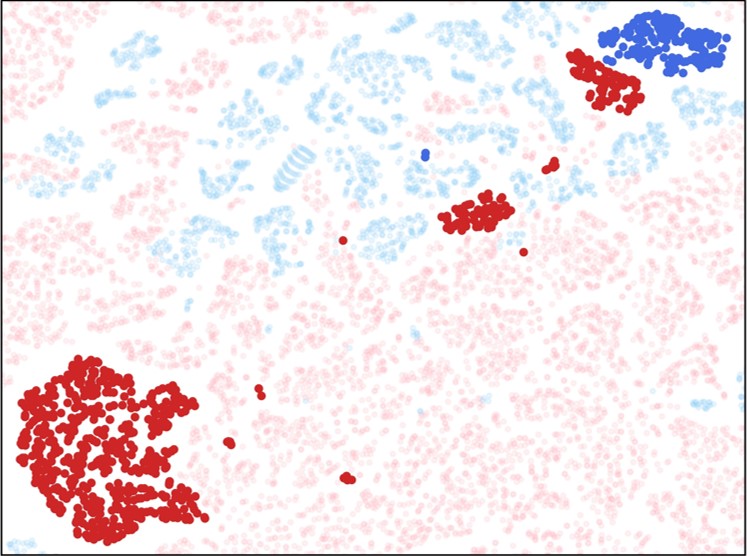}\\
(a) Images &  (b) Baseline & (c) Ours\\
\end{tabular}
\end{center}
\caption{
    Comparison of pixel distribution in feature space on the ADE20K validation set.
    Baseline: \textit{ResNet-101 + DeeplabV3+}.
    Ours: \textit{Baseline + CCS layer}.
    Pixels that randomly sampled form the ``tree" and ``plant" are colored in red and blue, respectively.
    Our distribution of each class is more compact than the baseline. There are also clear boundaries between different classes in the feature space of our method, when the pixels of the two classes are mixed up in the feature space of the baseline.
}
\vspace{-15pt}
\label{fig:comp}
\end{figure}

Due to the simplicity of the architecture of FCN,
existing methods mainly focus on enhancing the feature representations 
to improve visual recognition capability.
They aggregate rich contextual information via large receptive field~\cite{dilation, deformable}, 
multi-scale methods~\cite{psp, ASPP},
or attention mechanisms~\cite{attention, danet, psa, DUALA1, nonlocal}.
However, we observe that these methods still suffer from two challenges.
(i) \textbf{Large intra-class feature variation between different scenes}.
According to Fig.~\ref{samplefig}, the features of the ``trees" (colored in red) near the wall in \textit{Sample A} greatly differ from those on the bank in \textit{Sample B}, despite that they belong to the same semantic category.
Similarly, plant pixels of \textit{Sample A} locate significantly far from those of \textit{Sample B}, which also suggesting large intra-class variation between pixels in different scenes.
(ii) \textbf{Small inter-class distinction inside each scene}. 
Take \textit{Sample A} in Fig.~\ref{samplefig} for example, it is difficult for the segmentation network to tell pixels belonging to the ``tree" and pixels belonging to the ``plant" apart, since pixels belonging to the ``tree" are near to pixels of the ``plant", indicating that the two groups of pixels of different semantic classes bear similar features.
These two challenges make it hard to category pixels from the same class but different scenes into the correct class while telling groups of pixels from the different classes but the same scene apart from each other.

To solve the aforementioned challenges,
we rethink previous methods from a perspective of similarity between pixels and class centers.
Most previous methods utilize a $1\times1$ convolutional layer as the segmentation head to get the final prediction.
The weights of the convolutional layer are trained through all training samples and conduct a convolution operation on the feature maps in the inference phase.
In other words, the learned weights perform correlation calculation with each pixel and output the map of inner production which is widely used to measure the similarity between features.
As each vector of the weights can be regarded as a learned representation of the corresponding class, the final segmentation prediction map can also be regarded as a pixel-to-center similarity map.
These learned representations contain the common information of their class on the whole dataset.
Therefore, we call them global class centers.
In this view, the classification can be remodeled as follows: 
A pixel will be assigned to the class whose global class center is the most similar to the pixel among all classes. 
In conclusion, semantic segmentation can be viewed as a task to predict the similarity between pixels and class centers.

Motivated by this perspective,
we argue that there are two limitations of previous methods responsible for the above-mentioned challenges:
(i) The global class centers are incapable of 
adapting the large intra-class variations between different scenes, since the global class centers are learned based on the whole dataset and are kept unchanged and identical for different scenes during the inference stage.
(ii) Since there is no constraint on the similarities between global class centers,
these centers which represent different classes may share excessively similar representations,
leading to difficulties in correctly distinguishing pixels. 
These limitations correspond to the two challenges of the large intra-class variation between scenes and the small inter-class distinction inside each scene, respectively.

\begin{figure*}[htp]
\footnotesize
\centering
\renewcommand{\tabcolsep}{1pt} %
\renewcommand{\arraystretch}{1} %
\begin{center}
\begin{tabular}{cccccc}

\includegraphics[height=0.15\linewidth,width=0.20\linewidth]{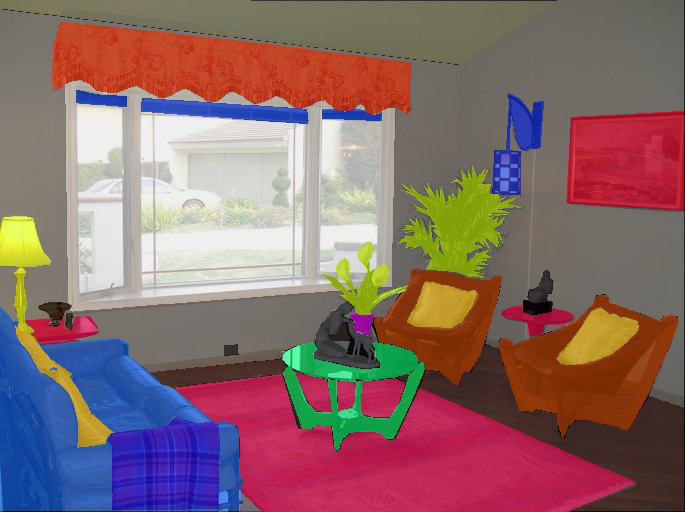} &
\includegraphics[height=0.15\linewidth,width=0.20\linewidth]{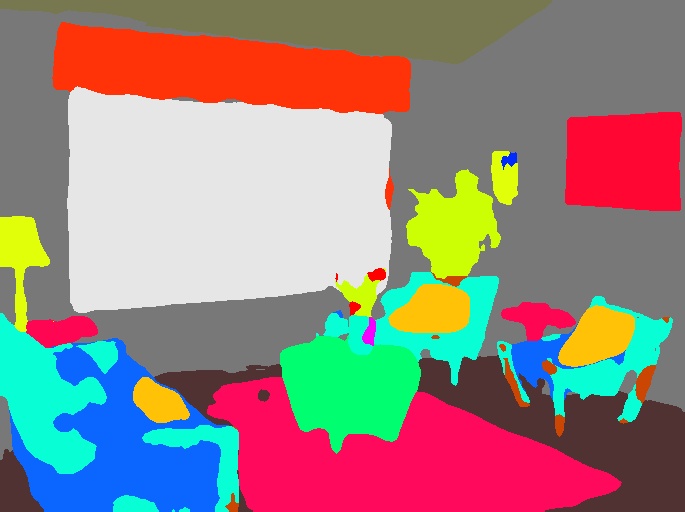}  &
\includegraphics[height=0.15\linewidth,width=0.20\linewidth]{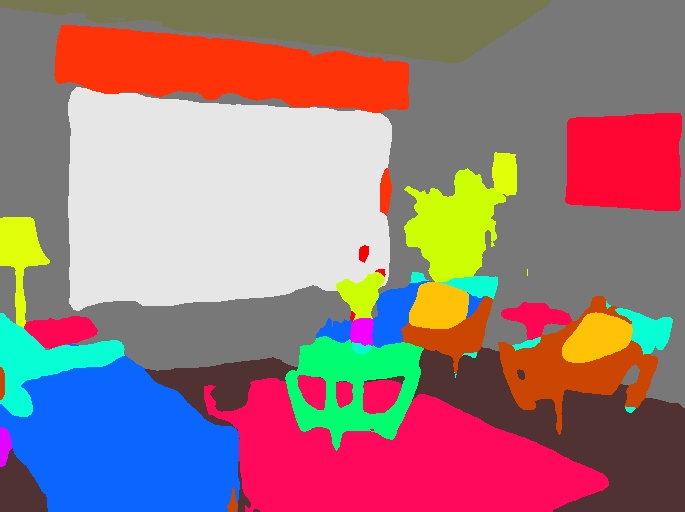} &
\includegraphics[height=0.15\linewidth]{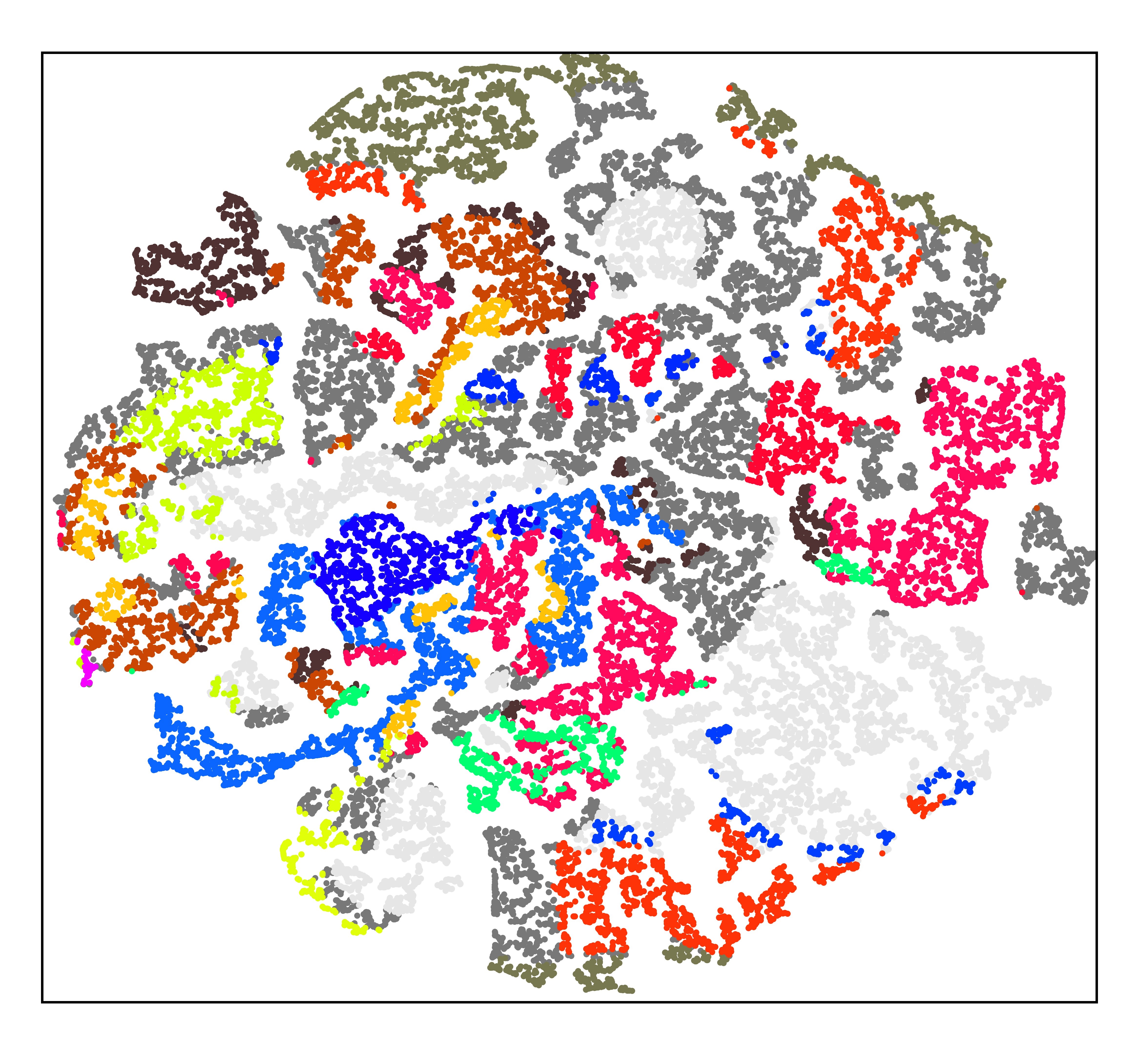} &
\includegraphics[height=0.15\linewidth]{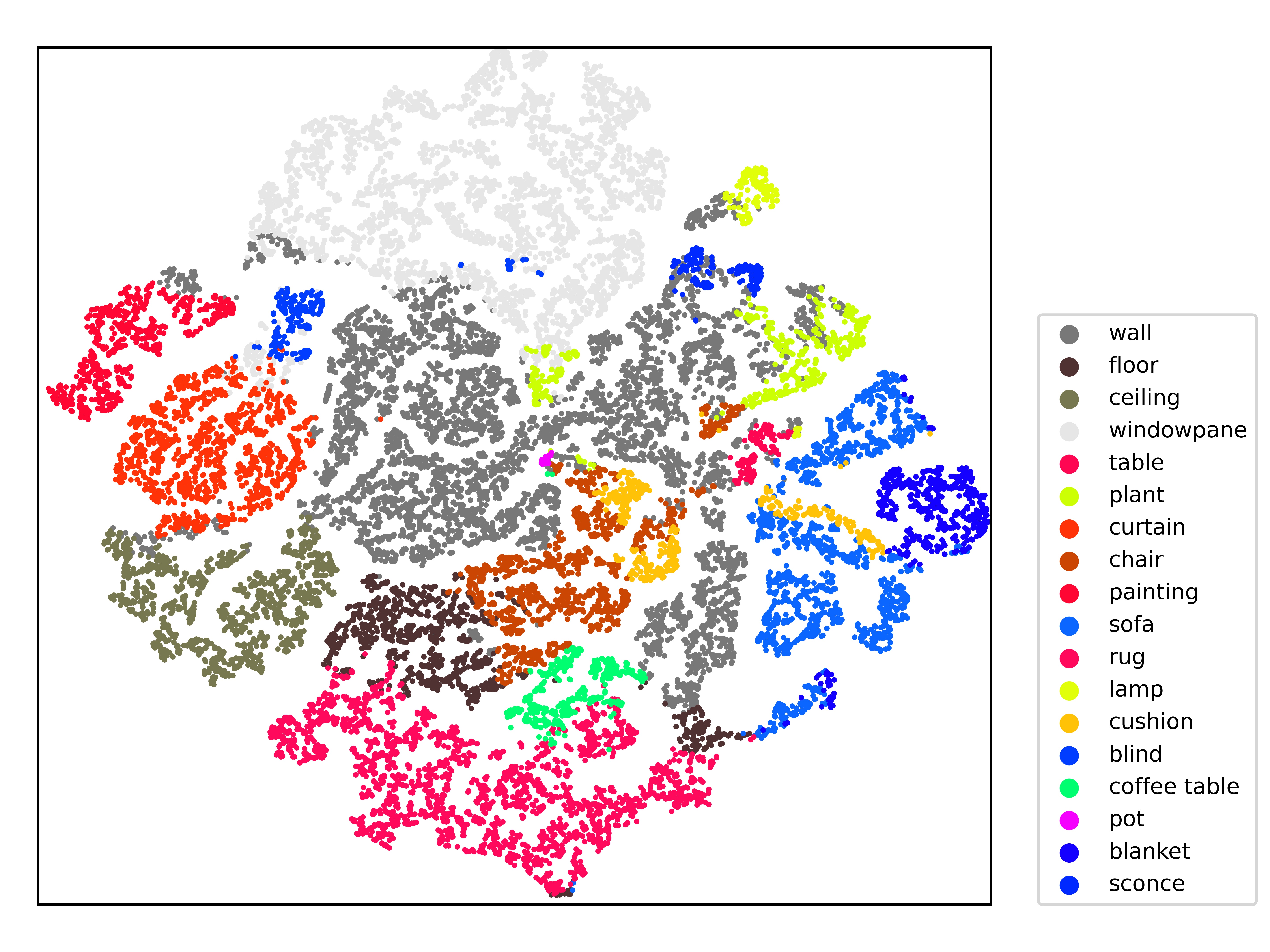} \\

\includegraphics[height=0.15\linewidth,width=0.20\linewidth]{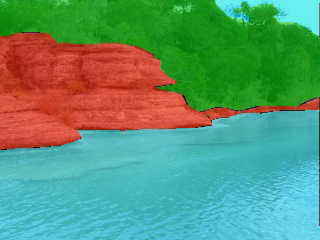} &
\includegraphics[height=0.15\linewidth,width=0.20\linewidth]{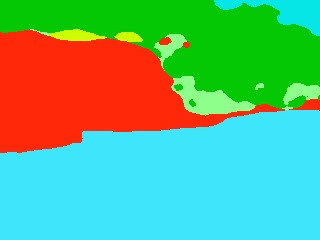}  &
\includegraphics[height=0.15\linewidth,width=0.20\linewidth]{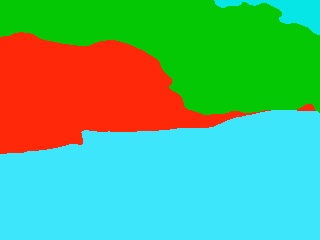}  &
\includegraphics[height=0.15\linewidth]{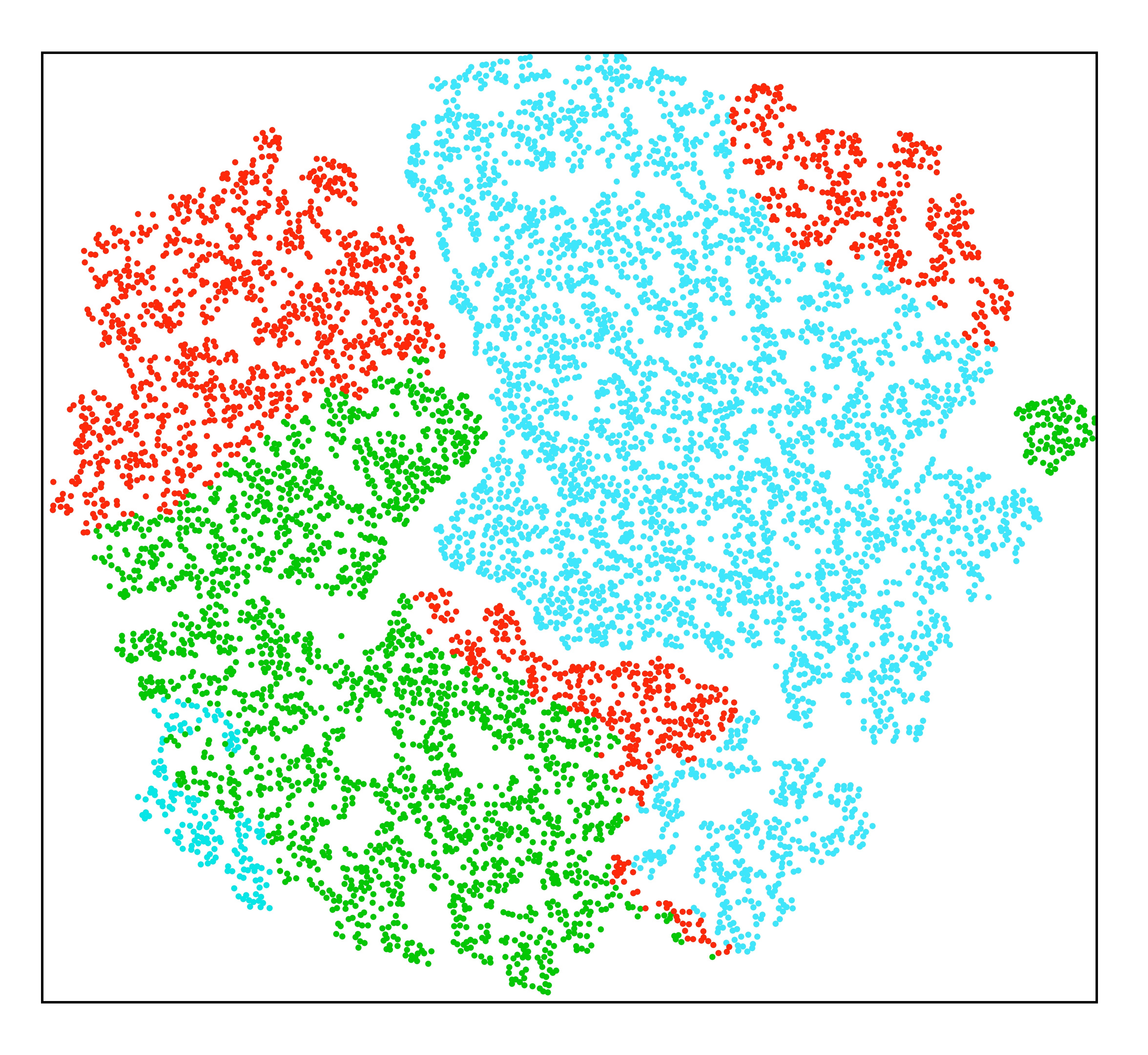} &
\includegraphics[height=0.15\linewidth]{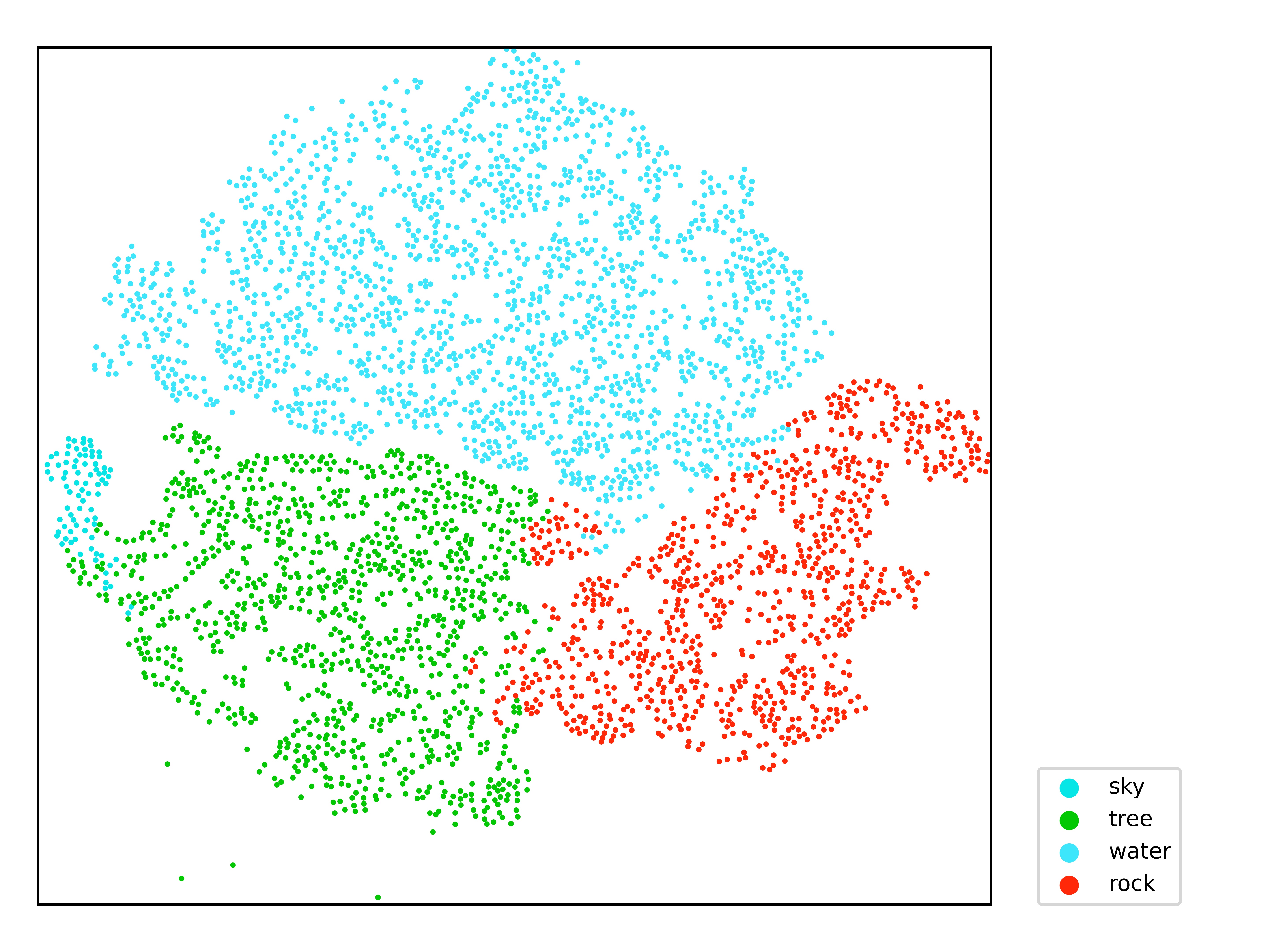} \\

\includegraphics[height=0.15\linewidth,width=0.20\linewidth]{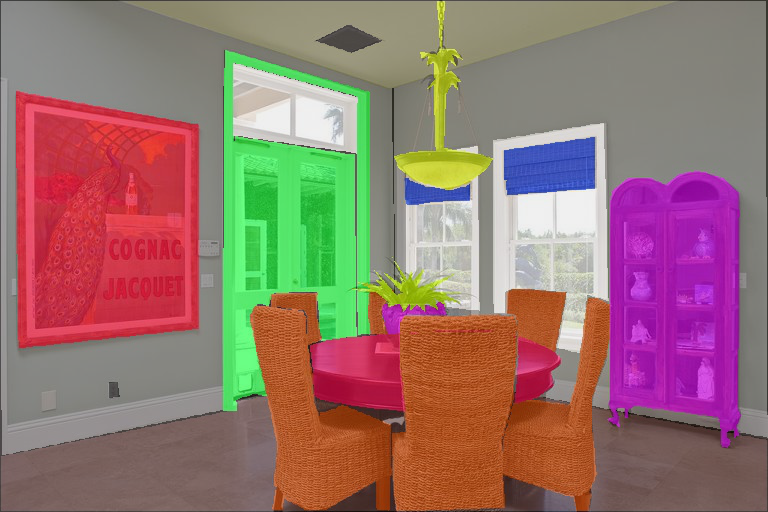} &
\includegraphics[height=0.15\linewidth,width=0.20\linewidth]{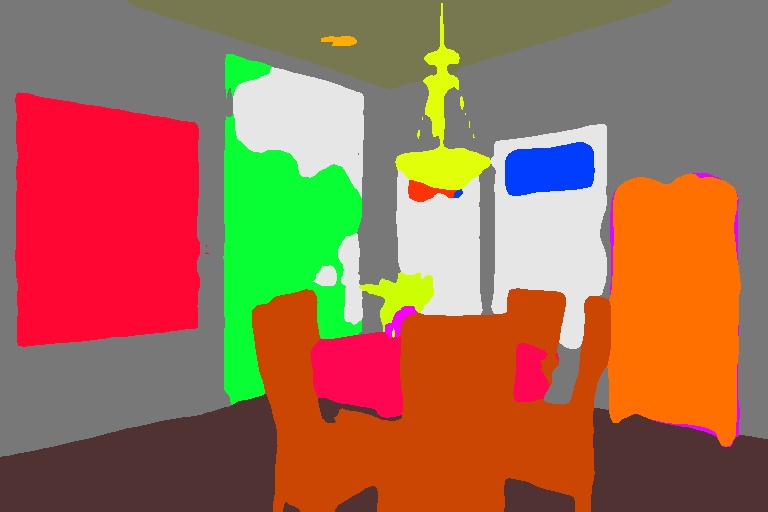}  &
\includegraphics[height=0.15\linewidth,width=0.20\linewidth]{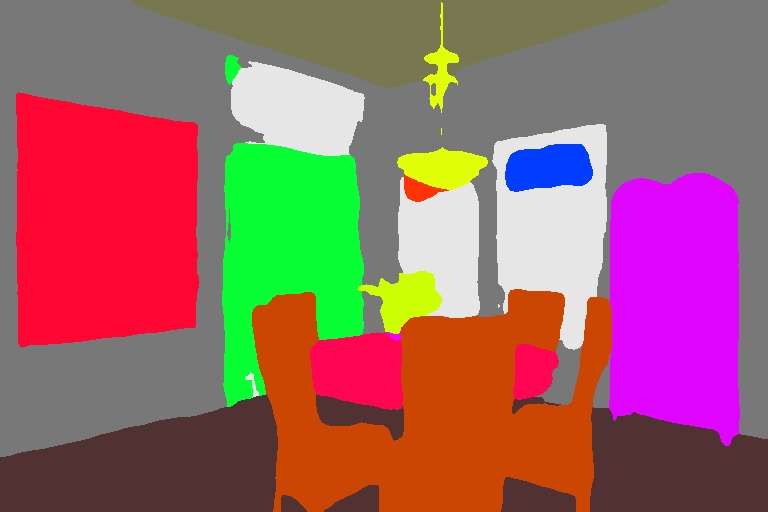} &
\includegraphics[height=0.15\linewidth]{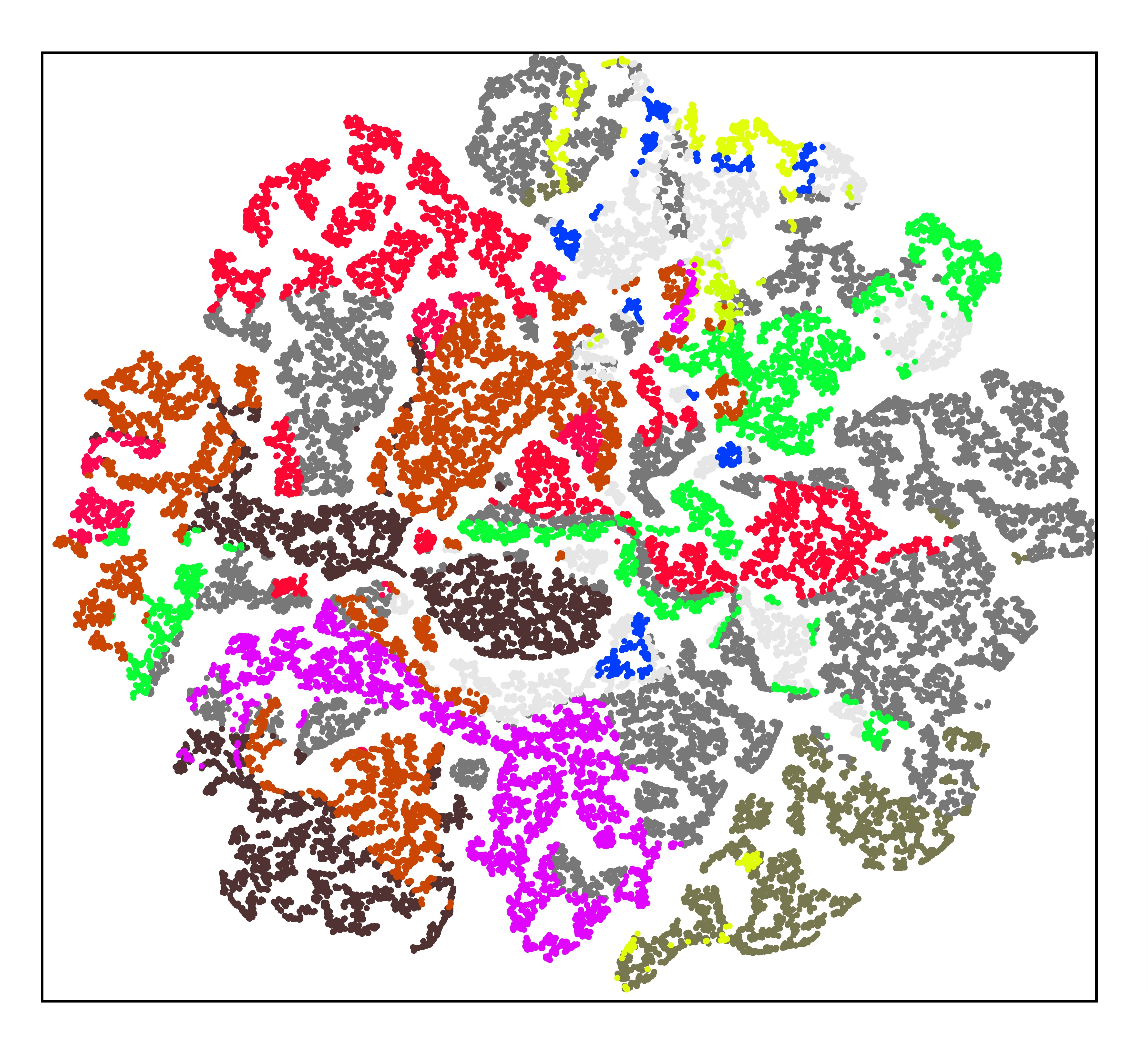} &
\includegraphics[height=0.15\linewidth]{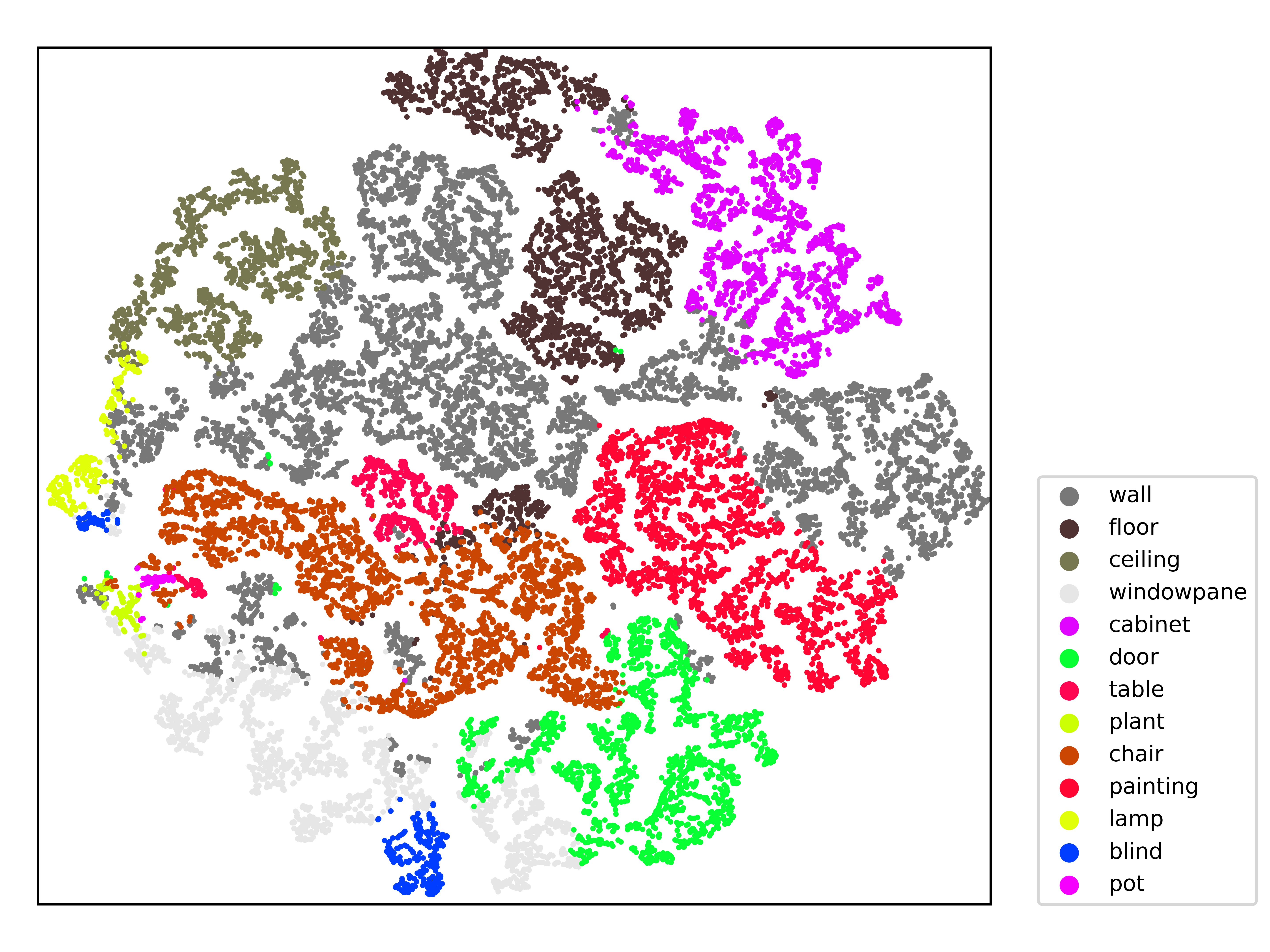} \\

\includegraphics[height=0.15\linewidth,width=0.20\linewidth]{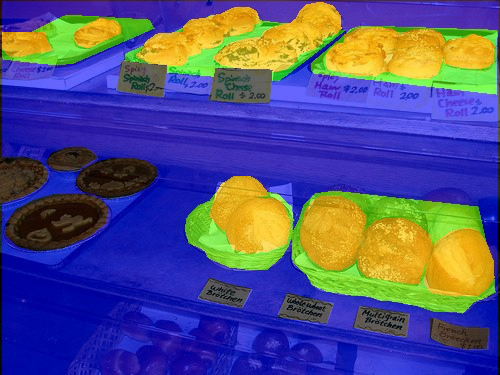} &
\includegraphics[height=0.15\linewidth,width=0.20\linewidth]{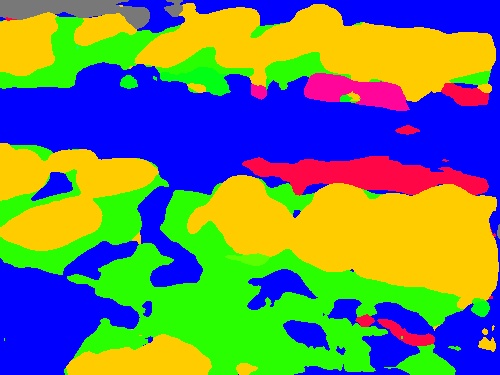}  &
\includegraphics[height=0.15\linewidth,width=0.20\linewidth]{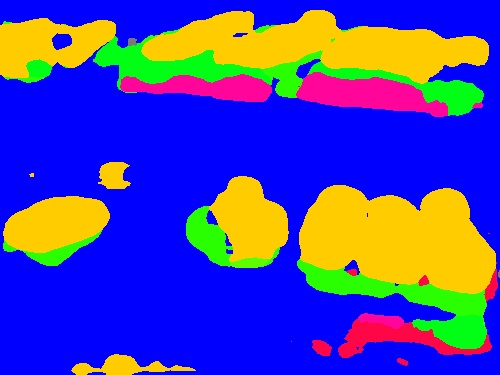}  &
\includegraphics[height=0.15\linewidth]{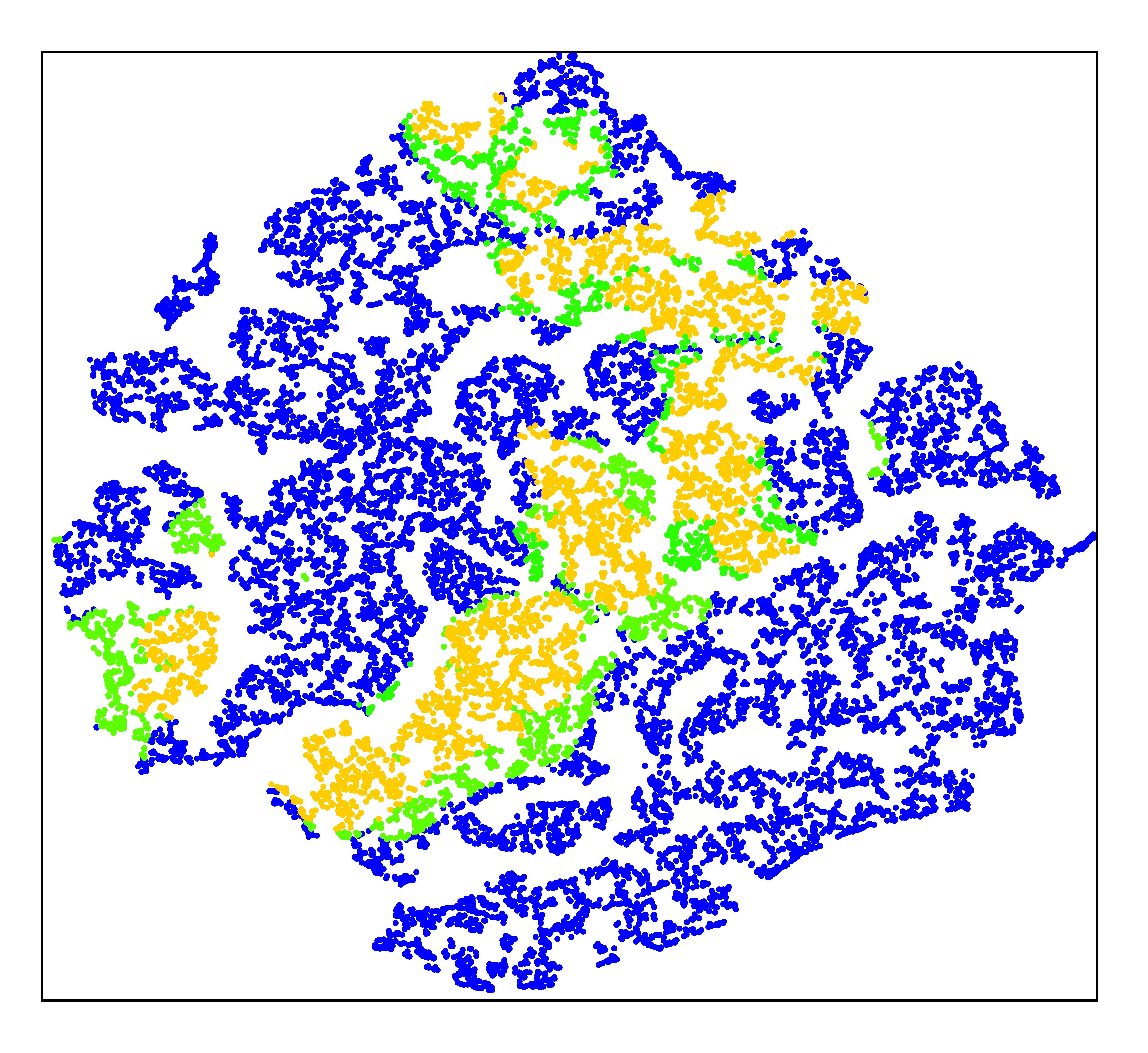} &
\includegraphics[height=0.15\linewidth]{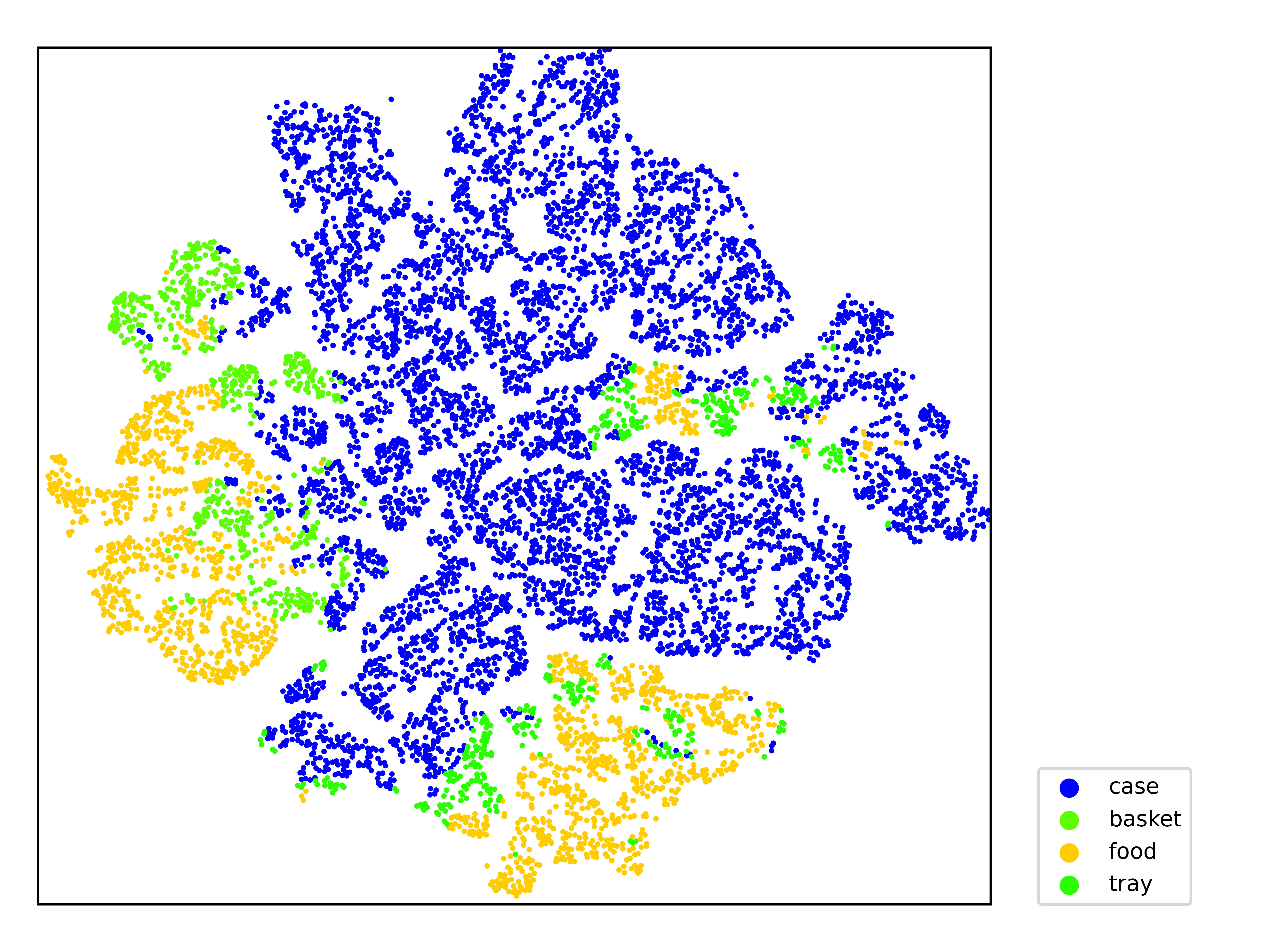} \\

\includegraphics[width=0.20\linewidth]{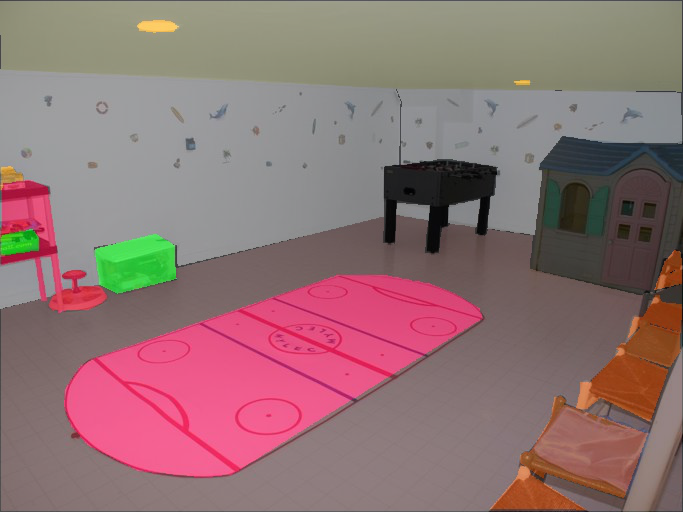} &
\includegraphics[width=0.20\linewidth]{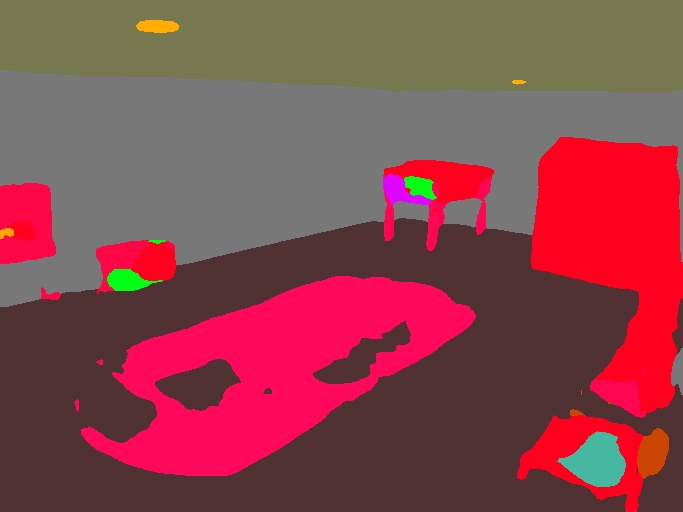}  &
\includegraphics[width=0.20\linewidth]{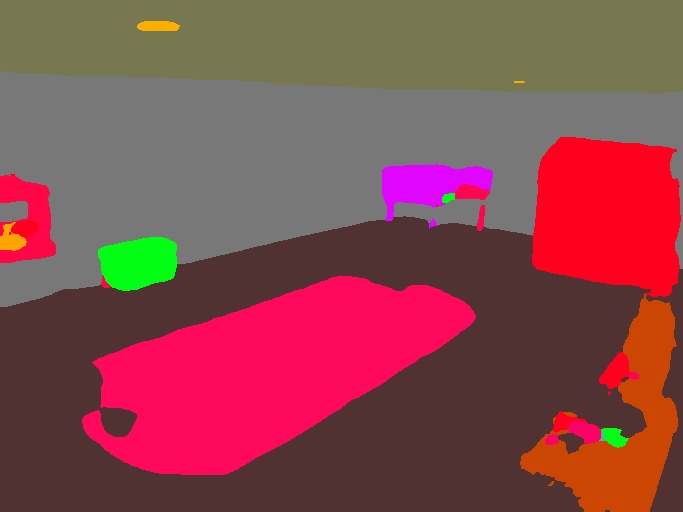}  &
\includegraphics[height=0.15\linewidth]{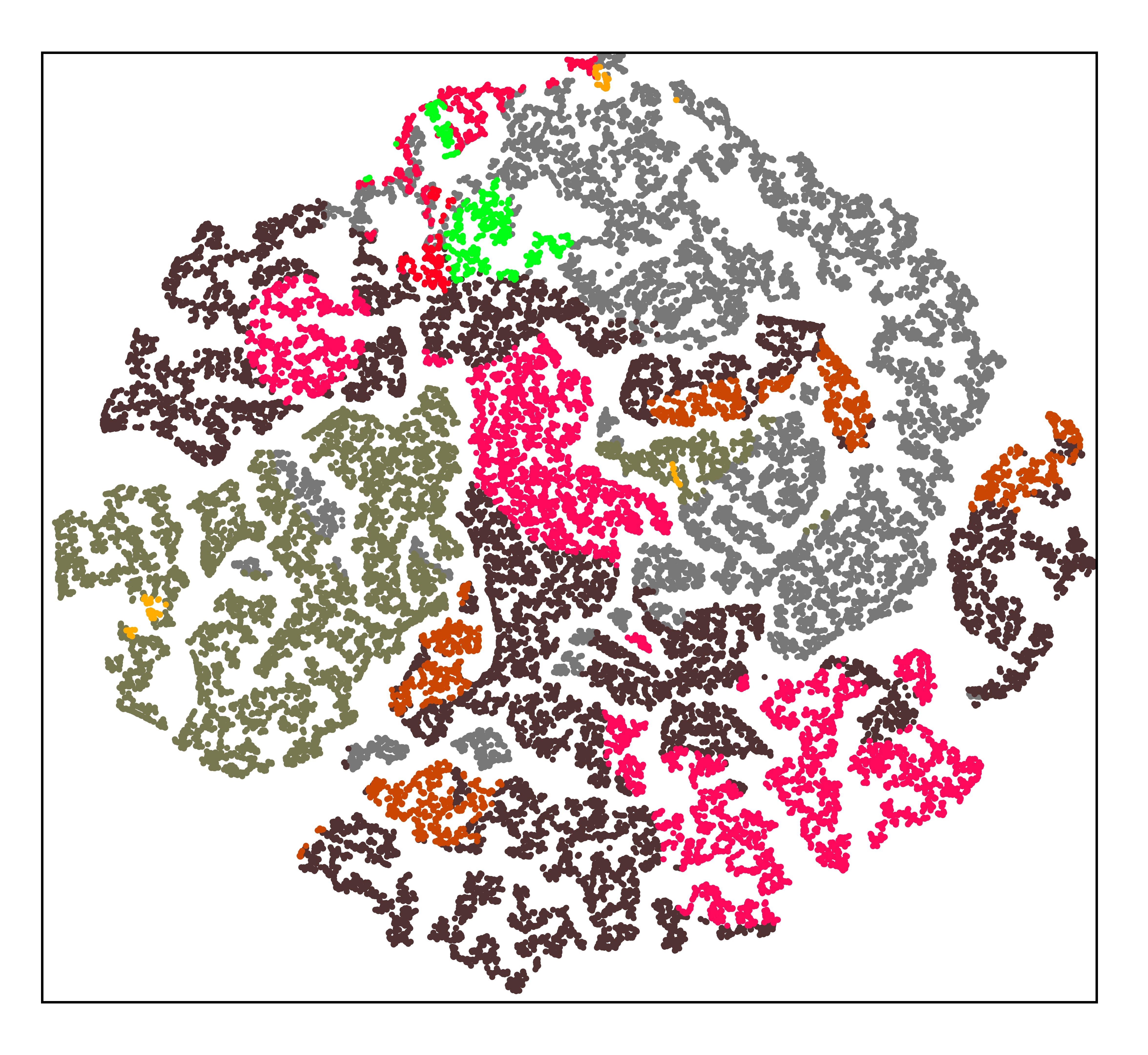} &
\includegraphics[height=0.15\linewidth]{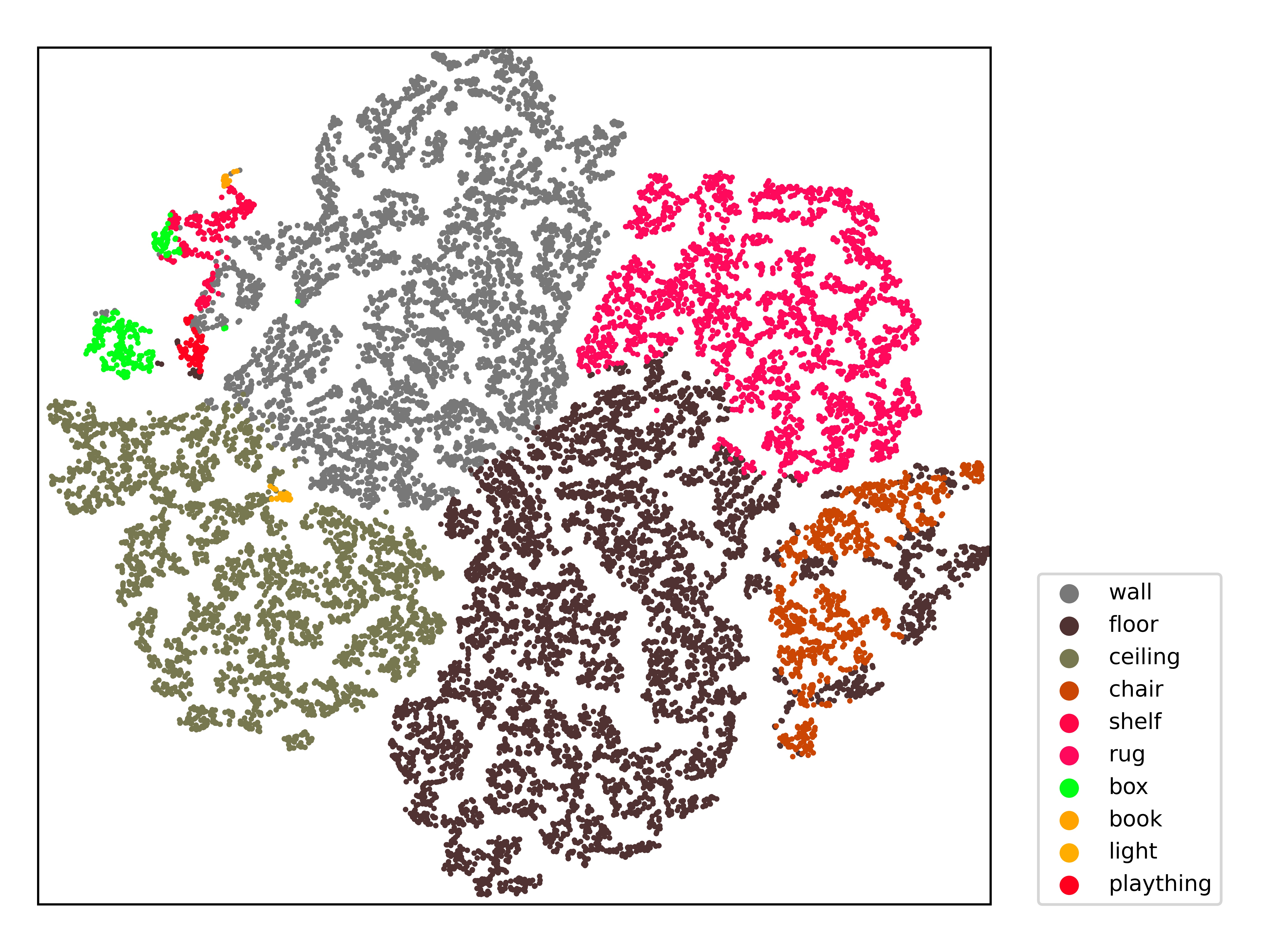} \\

\includegraphics[height=0.15\linewidth]{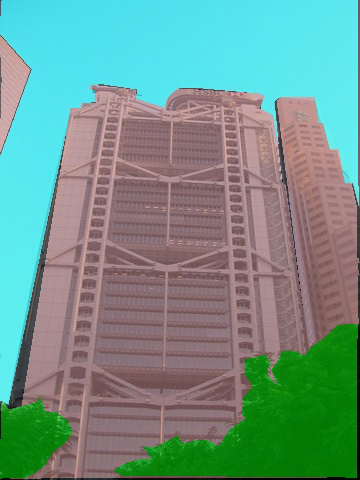} &
\includegraphics[height=0.15\linewidth]{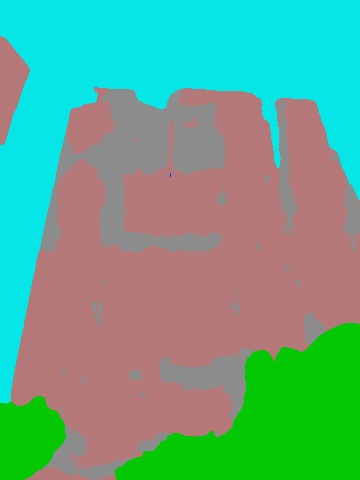}  &
\includegraphics[height=0.15\linewidth]{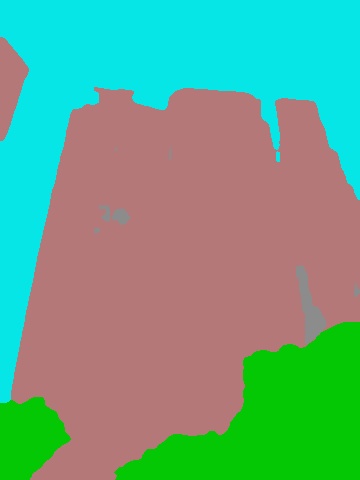}  &
\includegraphics[height=0.15\linewidth]{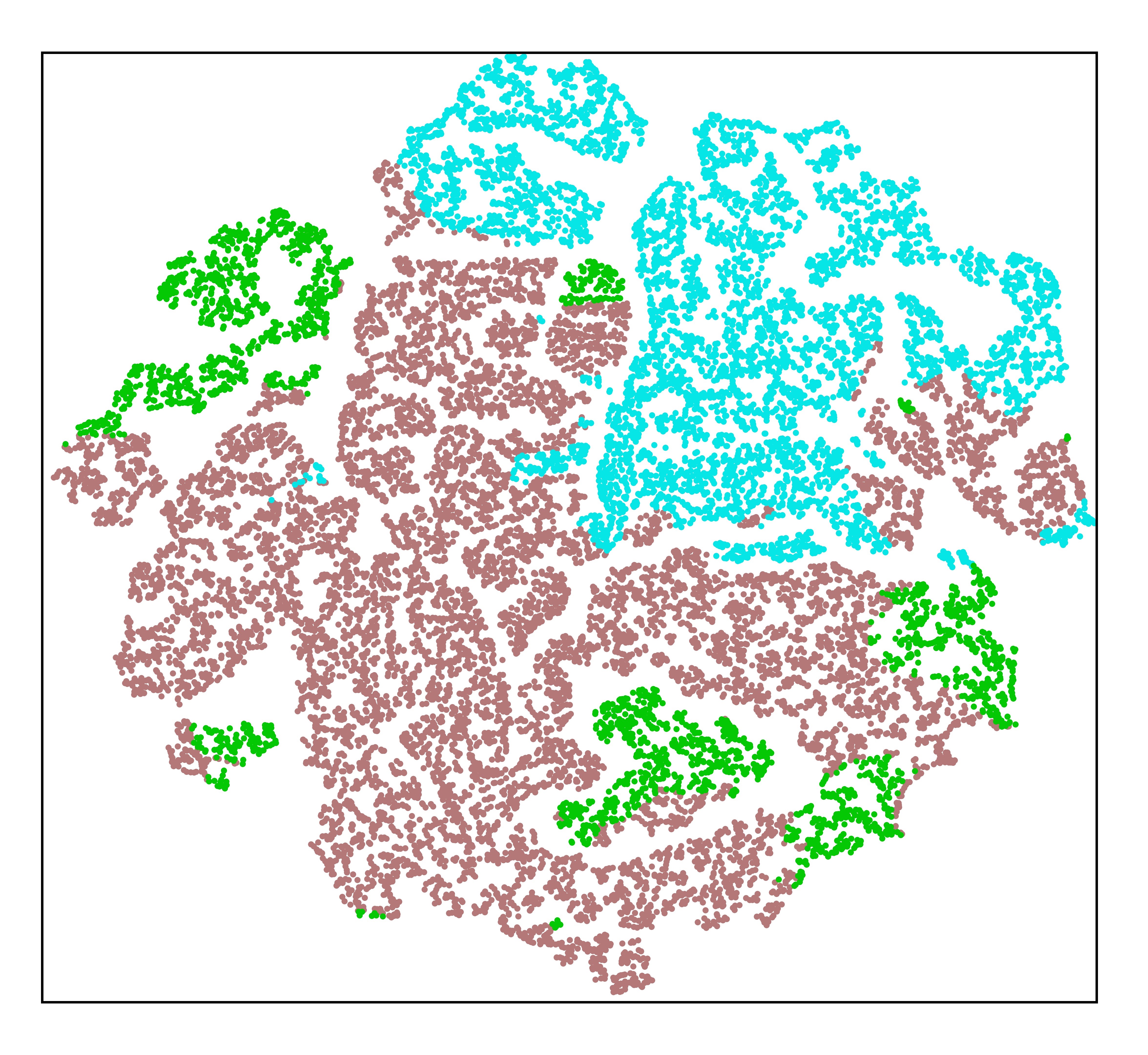} &
\includegraphics[height=0.15\linewidth]{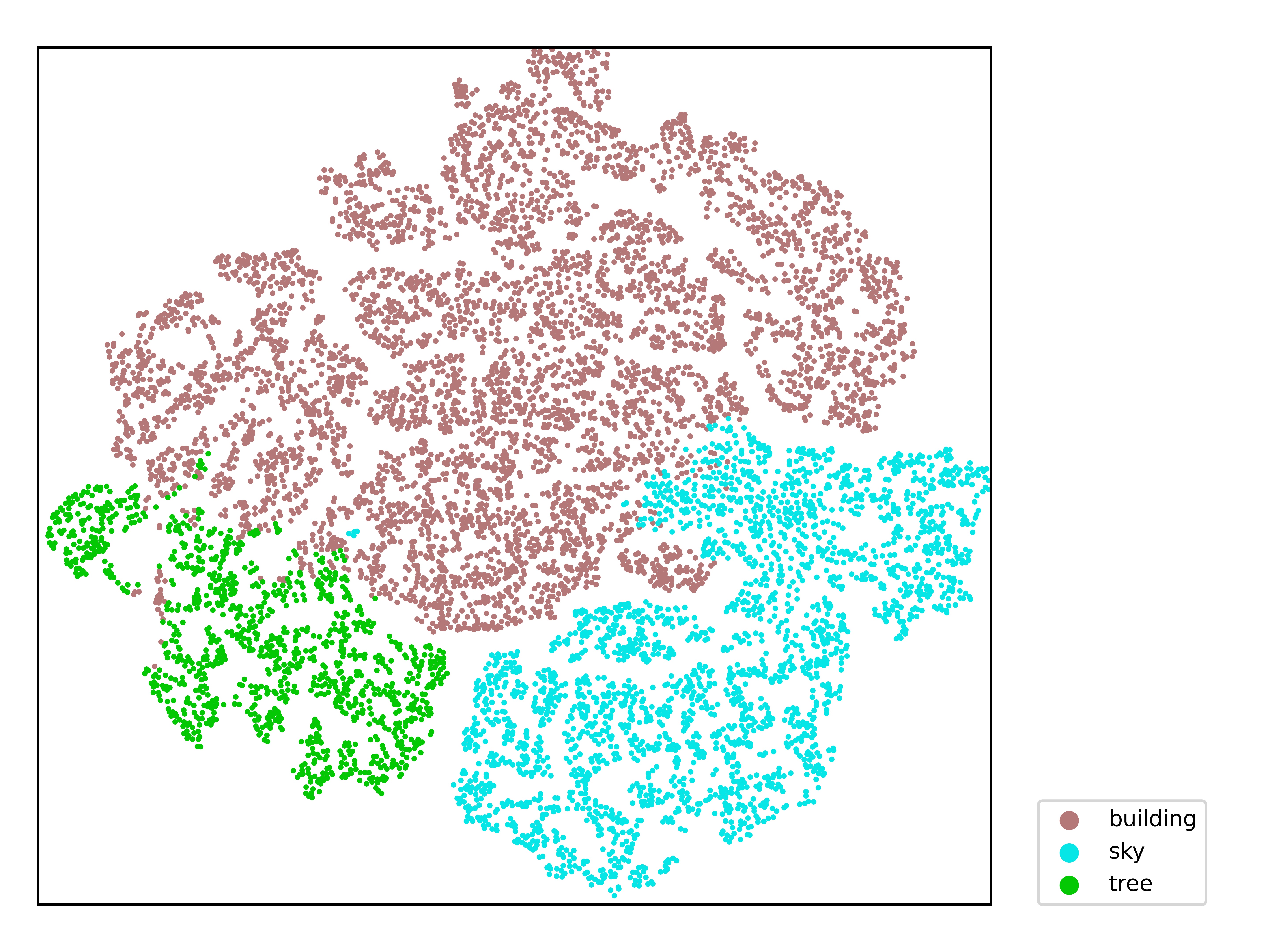} \\

\\

(a) Images &  (b) Baseline  & (c) Ours  & (d) Baseline distribution & (e) Our distribution\\
\end{tabular}
\end{center}
\caption{
    Visualization results on ADE20K validation set.
    We visualize the distribution results to demonstrate our model learns a better scene-level feature distribution than previous works. Comparing our method with the Baseline DeeplabV3+, we find that our boundaries between different clusters are far more clear, with most pixels of the same class located in the same cluster.
}
\label{fig:dist_spm}
\end{figure*}

To solve the challenges,
we propose a novel and flexible \textbf{Class Center Similarity}~(CCS) layer, which replaces the segmentation heads of networks and transfers the pixel-wise classification task to a pixel-to-center similarity prediction task.
Our CCS layer consists of three parts: Adaptive Class Center Module~(ACCM), Similarity Calculation Module~(SCM), and Class Distance (CD) Loss. 
First, Adaptive Class Center Module~(ACCM) generates adaptive class centers conditioned on each scene~(image), which accommodates the large intra-class variance between scenes.
Thus, the prediction only depends on the similarity between each pixel and the scene-specific adaptive class centers, instead of the immutable global class centers.
Then, the adaptive class centers are forward Similarity Calculation Module~(SCM) to compute the pixels-to-centers similarity between pixels and class centers and the mutual similarity of class centers as the inter-class similarity. 
Finally, our CD Loss is applied to the inter-class similarity and pixels-to-centers similarity in each scene to supervise the segmentation prediction while increasing the lack of inter-class distinction inside each scene simultaneously. 
The CCS layer can be integrated into 
almost arbitrary semantic segmentation architectures 
substituting for the final segmentation head, namely the $1\times1$ convolution layer. 
To demonstrate the effectiveness of the proposed Class Center Similarity layer, 
we add the CCS layer to existing segmentation networks and carry out extensive experiments on ADE20K and Pascal Context. Fig.~\ref{fig:comp} and Fig.~\ref{fig:dist_spm} also show that our method leads to clear and compact clusters for each semantic class in each scene.
In summary, the following \textbf{contributions} are made in this paper:
\begin{compactitem}

\item 
We rethink the semantic segmentation task %
from a \textit{pixel-to-center~(P-C) similarity} perspective. 
Semantic segmentation can be viewed as a task with two stages: compute P-C similarity for each pixel and categorize pixels to the most similar semantic class.
\item  
We propose a class Center Similarity layer that can generate conditional class centers for each scene under the constraint of our Class Distance Loss. Our CCS layer is easy to plug into almost any FCN-based semantic segmentation network.
\item 
We conduct extensive experiments to analyze the effectiveness of our approach. The proposed method, called CCSNet, achieves state-of-the-art performance on two challenging datasets. 
Based on ResNet-101~\cite{resnet}, our model achieves 47.76\% mIoU on ADE20K, and 54.9\% mIoU on PASCAL Context.
\end{compactitem}

\begin{figure*}[htpb]
\begin{center}
\includegraphics[width=0.9\linewidth]{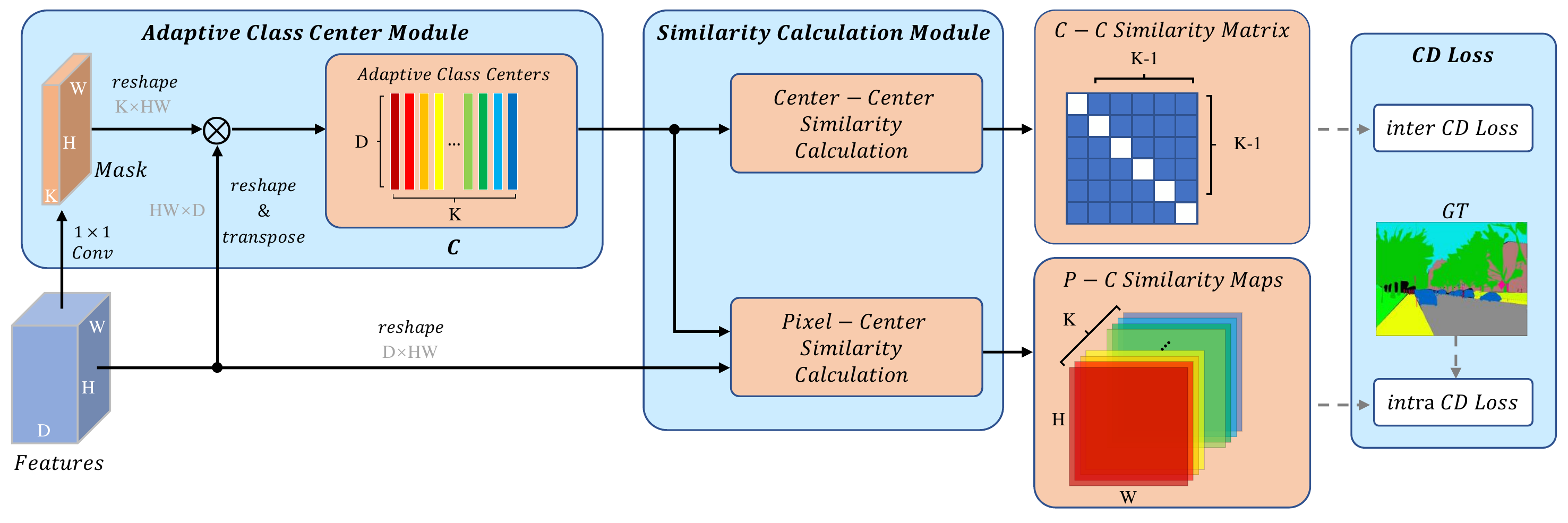}
\end{center}
\caption{Structure of the proposed CCS layer. 
The \textit{Mask} is a spatial weight map used for aggregating features for all the centers of each semantic class. 
$\mathbf{C}\in \mathit{R}^{K\times D}$ denote the generated $K$ adaptive class centers, where $D$ is the number of channels and $K$ is the number of classes.
Then, the CCS layer calculates the pixel-to-center~(P-C) similarity maps of each semantic class and center-to-center~(C-C) similarity between different classes. Finally, the proposed inter-class distance loss and intra-class distance loss are applied. The final prediction is P-C similarity Maps normalized by the softmax function.}
\label{fig:framework}
\end{figure*}

\section{Related Work}

\paragraph{Semantic segmentation.} The development of deep neural networks dramatically boosts semantic segmentation. Since FCN~\cite{long2015fully} replaces the fully connected layer in the traditional classification network with a convolutional to get pixel-wise predictions, FCNs achieves great success in semantic segmentation. Segnet~\cite{segnet}, UNet~\cite{unet} and RefineNet~\cite{refinenet} adopt encoder-decoder structure to recover the spatial information that lost by downsample operation via cascading upsampling. In order to capture long-range dependencies, lots of works have been done by introducing CRF~\cite{deeplab,crf2} and MRF~\cite{mrf} into segmentation tasks. Dilated convolution~\cite{dilation} and deformable convolution\cite{deformable}, which increase the resolution of feature maps, is used to enlarge the receptive field. We revisit these typical networks and find that semantic segmentation can be regarded as categorizing each pixel to the semantic category with the greatest pixel-to-center similarity (or the smallest pixel-to-center distance).

\paragraph{Contextual information.} Context is believed of great significance in semantic segmentation. Recently, plenty of works focusing on mining richer context information. Multi-scale representations are used to capture more context in PSPNet~\cite{psp}. the family of DeepLab~\cite{deeplab,ASPP,deeplabv3+} also captures the context information from multi-scales. Based on these approaches, many extensions have been proposed,\eg DenseASPP~\cite{denseaspp} and APCNet~\cite{apc}. The Attention mechanism is also adopted to capture global contextual information~\cite{huang2019ccnet, li2019global}. Other studies pay their attention to the similarity of pixels to help aggregate contextual information. OCNet~\cite{ocnet}, DANet~\cite{danet} and CFNet~\cite{cfnet} aggregate context that computed on all pixels and augment context representation to the representation of each pixel. Based on the self-attention mechanism, these approaches calculate similarity (or relation) between pixels and aggregate representations according to similarity.
Although a great number of studies explore discriminate representations to help segmentation, the large variance between pixels of the same category from different scenes and the lack of distinction in each scene still remains. 
Our work addresses these two challenges based on similarity as well, but we aim to get better similarity maps while those utilize context to get better features.
In addition, thanks to our ACCM which generates the varying adaptive class centers for classification, our method is more effective with minor computational cost.

\paragraph{Center-based methods in semantic segmentation.} Recently, several works propose approaches with centers in semantic segmentation networks. 
In contrast to previous works, ACFNet~\cite{zhang2019acfnet} presents the concept of class centers as the global context from a categorical perspective, which describes the overall representation of each class in a scene. Then, different class centers are adaptively concatenated with features according to each pixel for aggregating class-wise context. 
OCRNet~\cite{ocr} presents a simple method characterizing a pixel by exploiting the representation of the corresponding object class. Under supervision, OCRNet learns object regions and generate representations of object regions, which can be viewed as the object centers of each region. Pixel-region relation is computed to aggregate information from object centers. 
Similar to ACFNet, the object-contextual representations generated from object centers and pixel-region relation augment the original pixel representations via concatenation. 
Compared with these methods that also introduce the concept of center in their works, our concern is how to get conditional centers to address the drawback of global centers, while they follow the thought of exploring context from a different aspect. Besides, we apply our inter CD Loss and intra CD Loss on class centers to enlarge distinction and optimize the distribution of different classes while they do not have any supervision on the generated centers.

\section{Method}
In this section, we first have a review of typical segmentation models in semantic segmentation. Then, we explicate our rethinking-modeling semantic segmentation as a task to assign each pixel to a category directly based on the pixel-to-center (P-C) similarity.
After a brief overview of the proposed prediction pipeline,  
we introduce the details of the Adaptive Class Center Module, Similarity Calculation Module and Class Distance Loss.

\subsection{Revisiting prediction of semantic segmentation}
Before we go to our proposed approach, let's revisit the architecture of typical segmentation networks.
A typical segmentation network consists of two parts: \textit{feature extractor} 
and \textit{classifier}. The feature extractor is usually a deep convolutional network, which
takes images as inputs, then extracts high-dimensional representations of each pixel. 
The classifier takes the output features of the extractor as inputs and computes the score maps indicating the probabilities that every element belongs to each class. 
The output 
of the feature extractor is noted as $\mathbf{F}\in \mathcal{R}^{D\times N}$, 
where $N$ is the total number of elements, $D$ is the number of output channels in the extractor. The representation of the \emph{i-th} pixel in the input image is noted as $\mathbf{f}_i\in \mathcal{R}^{D\times 1}$. 
The weights of the classifier are noted as $\mathbf{W}^c\in \mathcal{R}^{K\times D}$, and
$\mathbf{w}_j\in \mathcal{R}^{1\times D}$ $(j=1,2,...,K)$ denotes the 
kernel weight which performs correlation on feature maps to get the prediction of \emph{j-th} class. 
Following the work of FCN, most segmentation networks replace the FC classifier with a $1\times 1$ convolutional layer, which computes the score map $\mathbf{S} \in \mathcal{R}^{K\times N}$ of input feature $\mathbf{F}$. The computation can be formulated as:
\begin{equation}
    \mathbf{S=W \circledast F} ,
\end{equation} where $\circledast$ is the convolution operation.
$\mathbf{S}$ is then normalized by soft-max to get a soft-margin prediction of the whole image.

As classifier employs a $1\times1$ conv layer, the computation of the \emph{j-th} class' score $s_{ij}$ of \emph{i-th} pixel can be formulated as:
\begin{equation}
    s_{ij}=\mathbf{w}^\top_j \cdot \mathbf{f}_i ,
    \label{eq:psimi}
\end{equation}
where $\cdot$ denotes the inner product, $s_{ij}$ is the score of \emph{i-th} pixel at \emph{j-th} class, and $\mathbf{w}^\top_j$ is the transposed $\mathbf{w}_j$, $i=1,2,...,N$.
The prediction $p_{i}$ assigned to \emph{i-th} pixel is made by performing a $argmax$ operation to each location in the image:
\begin{equation}
    p_{i}= \mathop{\arg\max}_{j} (s_{ij})= \mathop{\arg\max}_{j} (\mathbf{w}^\top_{j}\cdot\mathbf{f}_i) .
\end{equation}

Hence, for a single pixel, the score $s_{ij}$ is actually the inner production similarity, or so-called correlation, between its feature $\mathbf{f}_i$ and the transposed weight vector $\mathbf{w}^\top_j$ of class $j$. 
According to the pixel-to-weight similarity map $\mathbf{S}$, the final prediction is made by categorizing pixels to the semantic class which has the greatest similarity value in feature space. 
Thus, the weight vectors can be deemed as learned class centers, and the category assignment is equivalent to \textit{finding the nearest class center for each pixel}. 
Consequently, the overall inference procedure can be summarized as Alg.~\ref{alg prev}, where $\mathbf{I}\in \mathcal{R}^{3\times N}$ is the input RGB features of an image, $\theta$ is the parameter of network, $\mathbf{W}^c \in \mathcal{R}^{K\times D}$ is the weights of the classifier, namely the learned global class centers, $i=1,2,...,N$, $j=1,2,...,K$.

\begin{algorithm}[!ht]
\caption{Inference with Global Class Centers}
\label{alg prev}
\begin{algorithmic}
\renewcommand{\algorithmicrequire}{\textbf{Input:}}
\renewcommand{\algorithmicensure}{\textbf{Output:}}
\REQUIRE  $\mathbf{I}$, network $Net(,)$ and parameters $\mathbf{\theta}$, weights of classifier $\mathbf{W}^c$.
\vspace{0.08cm}
\STATE Extract feature map:  $\mathbf{F} \leftarrow  Net(\mathbf{I},\mathbf{\theta})$.
\vspace{0.05cm}
\STATE Calculate P-C similarity:  $\mathbf{S} \leftarrow Simi(\mathbf{F},~ \mathbf{W}^c)$
\STATE Get prediction: $p_{i} \leftarrow \mathop{\arg\max}_{j} (s_{ij})$
\vspace{0.02cm}
\ENSURE Final prediction $\mathbf{P}$.
\end{algorithmic}
\end{algorithm}
\vspace{-0.5cm}
\begin{algorithm}[!ht]
\caption{Inference with Adaptive Class Centers}
\label{alg ours}
\begin{algorithmic}
\renewcommand{\algorithmicrequire}{\textbf{Input:}}
\renewcommand{\algorithmicensure}{\textbf{Output:}}
\REQUIRE  $\mathbf{I}$, network $Net(,)$, the parameters $\mathbf{\theta}$, and the ACCM $\mathcal{G}(\cdot)$.
\vspace{0.08cm}
\STATE Extract feature map:  $\mathbf{F} \leftarrow  Net(\mathbf{I},\mathbf{\theta})$.
\vspace{0.05cm}
\STATE Gnerate Adaptive Class Centers:  $\mathbf{C} \leftarrow  \mathcal{G}(\mathbf{F})$.
\vspace{0.05cm}
\STATE Calculate P-C similarity:  $\mathbf{S} \leftarrow Simi(\mathbf{F},~ \mathbf{C})$
\STATE Get prediction: $p_{i} \leftarrow \mathop{\arg\max}_{j} (s_{ij})$
\vspace{0.02cm}
\ENSURE Final prediction $\mathbf{P}$.
\end{algorithmic}
\end{algorithm}

If pixel $i$ belongs to the \emph{j-th} class according to ground truth, we hope that we can get greater relative $s_{ij}$ at the \emph{j-th} channel compared with other channels and smaller scores at irrelevant channels, which correspond to irrelevant classes. 
So, the network is trained under the supervision of Cross-entropy~(CE) Loss that applied on the normalized scores maps:
\begin{eqnarray}
    \mathcal{L}_{CE}&=&
    \sum_{i} -\mathit{log}\bigg( \frac{\mathit{exp}(\mathbf{s}_i\cdot\mathbf{y}_i)}{\sum_{j }\mathit{exp}(s_{ij})}\bigg) \nonumber \\
    &=&\sum_{i}-(\mathbf{s_i\cdot\mathbf{y}_i})+\mathit{log}\big(\sum_{j }\mathit{exp}(s_{ij})\big)  ,
    \label{CE}
\end{eqnarray}
where $\mathbf{y}_i$ is the one-hot label of element $i$.

In our understanding, CE Loss mainly focuses on enlarging the relative inner production similarity between the corresponding weight vector and pixels of each class by pushing pixels and corresponding weight vectors closer in feature space. 
Accordingly, we can regard the weight vectors as a set of global class centers learned by the network on the whole dataset.
The training stage is a process that the network learns to find the global class centers of each class, best fitting the distribution of all the pixels which belong to the same class on the whole training set.
Moreover, these global class centers are identical for all input scenes.

However, semantic segmentation networks are still facing two challenges: 
(i) pixels of the same category yet different scenes are significantly different in feature space, leading to difficulty for networks in categorizing all these pixels to the same semantic class.
(ii) pixels from different categories but the same scene lack enough distinction, causing trouble finding distinctive separating plane to tell pixels belonging to different classes apart.

To demonstrate the challenges, we conduct experiments and visualize the feature representations, which are the inputs of the final convolution, in Fig.~\ref{samplefig} of randomly sampled pixels
from ``tree" and ``plant" classes on ADE20K validation set, with sampling ratio set as 1\%. 
We adopt t-SNE~\cite{tsne} for a clear visualization.
As shown in Fig.~\ref{samplefig}, the pixels of ``trees" and ``plants" in \textit{Sample A} are far from those in \textit{Sample B} in feature space despite the fact that they belong to the same class.
This phenomenon indicates the large variance of features between different scenes, making it hard to assign pixels of the same class yet different scenes with the same class based on the immutable global class centers.
Moreover, for those pixels in the same scene, they are much closer to each other despite that they belong to different categories.
These short distances in feature space between pixels from the same scene but different classes corroborate the scarcity of enough distinction to categorize them correctly. 

\subsection{Overview of our proposed method}
As illustrated above, considering the learned class centers $\mathbf{W}^c$ are unchangeable for different input images $\mathbf{I}$, the first challenges are unavoidable.
To solve it, we propose a new pipeline for semantic segmentation prediction pipeline as shown in Alg.~\ref{alg ours}. 
The pixel-to-center~(P-C) similarity is calculated between pixels in each scene and the adaptive class centers which are generated based on the feature map of the scene.
Compared with $\mathbf{W}^c$ which is immutable during inference for every input image, the adaptive class center should be capable of varying between scenes to accommodate the large intra-class feature variation
in different scenes.

On the other hand, there is no constraint on the similarity between different class centers inside each scene.
So, the second challenge is aggravated, even though the adaptive class centers can mitigate the problem.
Therefore, we propose the Class Distance~(CD) Loss to enlarge the inter-class feature variation in the same scene.
CD Loss directly requires large similarities between pixels and the adaptive class center of the ground truth class and small mutual similarities among adaptive class centers.
Extensive experiments and ablation studies in Sec.~\ref{sec:exp} corroborate the effectiveness of our proposed prediction pipeline.

\subsection{Adaptive Class Center Module}
Previous approaches take the transposed learned weight $\mathbf{W}^c$ of the final classifier as global class centers, which are the representations of each class at the dataset-level. On account of the feature variances on the dataset from scene to scene, these approaches are impeded by their global class centers which are not able to vary between scenes. Therefore, we propose the Adaptive Class Center Module (ACCM) to generate unique class centers for each scene as class representations at the scene-level instead of the dataset-level.

ACCM performs matrix multiplication of pixel features and a learned mask to generate coarse class centers for each input image. 
This module refines the coarse class centers and outputs a set of adaptive centers as scene-level representations for all classes.
The whole computation process of ACCM can be formulated as:
\begin{equation}
    \mathbf{C}= \mathcal{A}(\mathbf{M} \otimes \mathbf{F}^\top), \label{cccm}
\end{equation}
where $\mathbf{C} \in \mathcal{R}^{K\times D}$ is the matrix of generated conditional adaptive class centers, $\otimes$ is matrix multiplication, $\mathcal{A}(\cdot)$ is the adaptive module consisting of several convolutional layers, and $\mathbf{M}\in \mathcal{R}^{K\times N}$ is a learned weight map, based on which ACCM aggregates information and generates the coarse class centers. 
In practice, we find that for the adaptive module, a $1\times1$ convolution works well to generate the adaptive class centers. We also conduct experiments to study the impact of different $\mathbf{M}$ and eventually choose the one supervised by Dice loss~\cite{dice} for better performance.
The Dice loss is defined as:
\begin{equation}
    \mathcal{L}_{Dice} = 1- \frac{2\sum^N_{i}\mathbf{m}_i\cdot\mathbf{y}_{i}}{\sum^N_{i}||\mathbf{m}_i||^2+\sum^N_{i}||\mathbf{y}_i||^2 + \epsilon},
\end{equation} where $\mathbf{m}_i\in\mathit{R}^{K\times 1}$ is the \emph{i-th} column vector in $\mathbf{M}$ corresponding to \emph{i-th} pixel, $||\mathbf{a}||$ is the second norm of vector $\mathbf{a}$, and $\epsilon$ is set as $1e^{-3}$ to prevent division by zero.

\begin{figure}[htbp]
\centering
\subfigure[traditional centers]{\includegraphics[width=0.45\linewidth]{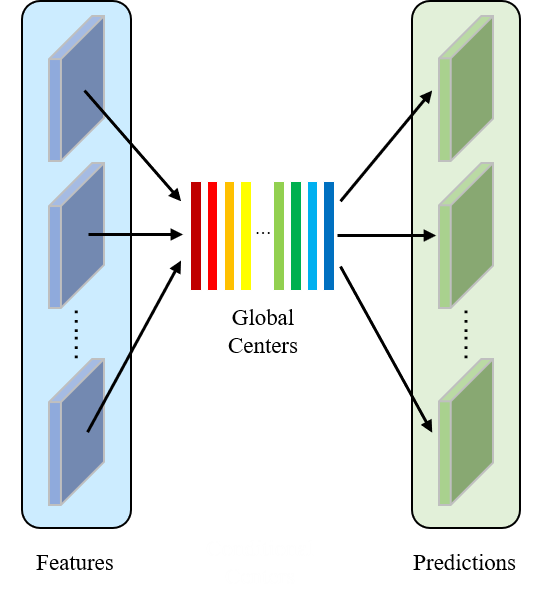}}
\subfigure[our centers]{\includegraphics[width=0.45\linewidth]{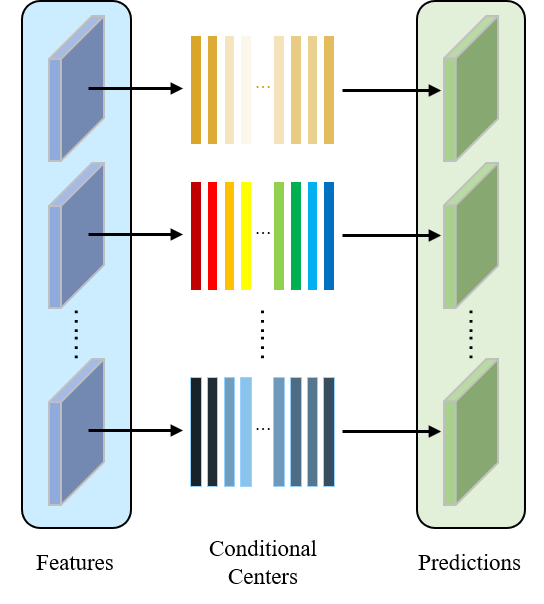}}
\caption{Illustration of our adaptive class centers. During inference, traditional approaches calculate pixel-to-centers similarity maps with fixed centers, while our approach generates adaptive centers for each input image, respectively.}
\vspace{-5pt}
\label{fig:condcenter}
\end{figure}

Based on the adaptive class centers, we perform pixel-to-center similarity calculation on feature maps to get absolute P-C similarity maps, followed by soft-max at the dimension of different classes to generate the relative P-C similarity maps as the soft-margin prediction. The structure of the model is shown in Fig.~\ref{fig:framework}.

Moreover, to address the problem that features of different classes but the same scene lack enough distinction, previous methods try to aggregate more context information into pixel representations, but there is no supervision imposed on the representations of class centers. We propose the Class Distance Loss composed of inter-class distance loss and intra-class distance loss imposed on class centers to make features in the same scene more discriminative. We will introduce it in the next section.

\subsection{Similarity Calculation Module}

We introduce of similarity function to measure the proximity of two embedding $\mathbf{a}\in \mathcal{R}^{D\times 1}$ and $\mathbf{b}\in \mathcal{R}^{D\times 1}$ in the feature space.
we define the inner production similarity as:
\begin{eqnarray}
    Simi(\mathbf{a}, \mathbf{b})_{inn} =\mathbf{a}\cdot\mathbf{b}.
    \label{sip}
\end{eqnarray}

To encourage class centers to be linearly independent and obviate the influence of the embedding's norm, we also introduce the cosine similarity as :
\begin{eqnarray}
    Simi(\mathbf{a}, \mathbf{b})_{cos} =\frac{abs(\mathbf{a}\cdot\mathbf{b})}{||\mathbf{a}||~||\mathbf{b}||},
    \label{scos}
\end{eqnarray}
where the $abs(\cdot)$ denotes the absolute value.

To calculate the C-C similarity, we employ consine similarity~(Eq.~\ref{scos}) to get rid of the influence of the norms of class centers. We simply replace $\mathbf{a}$ and $\mathbf{b}$ in Eq.~\ref{scos} with adaptive class centers $\mathbf{c}^\top_p$ and $\mathbf{c}^\top_p$ as $Simi(\mathbf{c}^\top_p, \mathbf{c^\top_q})$, where $\mathbf{c}_q \in \mathit{R}^{1\times D}$ is the \textit{q-th} row in $\mathbf{C}$, namely the adaptive class center of the \textit{q-th} class.

For P-C similarity calculation, we employ inner production simialrity~(Eq.~\ref{sip}). 
As the prediction is made by the argmax operation on P-C similarity maps, the absolute value of similarity between a pixel with a class center alone is meaningless unless compared with similarities between the pixel and other class centers. Hence, we introduce relative similarity as the probability that \emph{i-th} pixel belongs to \emph{q-th} class:
\begin{eqnarray}
    RSimi(\mathbf{f}_i, \mathbf{c}^\top_q) =\frac{\mathit{exp}\big(Simi(\mathbf{f}_i, \mathbf{c}^\top_q)\big)}{\sum_{j }\mathit{exp}\big(Simi(\mathbf{f}_i, \mathbf{c}^\top_j)\big)},
    \label{rsimi}
\end{eqnarray}
where $j=1,2,...,K$.

\subsection{Class Distance Loss}
Since there is no constraint on the similarity between different class centers inside each scene, the challenge that different-classes pixels in the same scene lack enough distinction is aggravated consequently.
We believe that it is easy for pixel-wise classification if pixel representations in the same scene have large inter-class distinction and small intra-class distinction. 
Motivated by this notion, we propose the Class Distance~(CD) Loss.
We introduce the inter-class distance and intra-class distance at first, then explicate the details of the Class Distance Loss.

\paragraph{Definition of inter-class and intra-class distance.}
The inter-class distance between the \emph{p-th} and the \emph{q-th} class is defined as:
\begin{eqnarray}
    d^{(p,q)}_{inter} = D(\mathbf{c}^\top_p, \mathbf{c}^\top_q),
\end{eqnarray}
where $p$ and $q$ are the numbers of two different classes, $\mathbf{c}^\top_p$ and $\mathbf{c}^\top_q$ are the class center vectors of \emph{p-th} and \emph{q-th} class respectively, $D(\cdot, \cdot)$ is the distance function.
The intra-class distance of the \emph{q-th} class is defined as:
\begin{eqnarray}
    d^q_{intra} = \sum^{N}_{i=1}\mathbbm{1}[y_{iq}=1] D(\mathbf{f}_i, \mathbf{c}^\top_q), \label{dintra}\\
    \mathbbm{1}[condition]=
\begin{cases}
1& condition~is~True\\
0& condition~is~False
\end{cases},
\end{eqnarray}

We introduce our distance function based on the relative similarity. 
The distance should be negative correlated with relative similarity. Therefore, we define our distance as: 
\begin{eqnarray}
    D(\mathbf{f}_i,\mathbf{c}^\top_q)= -log\big( RSimi ( \mathbf{f}_i,\mathbf{c}^\top_q) \big ).
    \label{dlog}
\end{eqnarray}
We apply inner production similarity and cosine similarity for $d_{intra}$ and $d_{inter}$, respectively.

\paragraph{Inter-class and intra-class Distance Loss.}
As illustrated above, we propose inter-class distance loss and intra-class distance loss to enlarge inter-class distances and diminish intra-class distance.
Our loss functions are as:
\begin{eqnarray}
&&  \mathcal{L}_{intra} = \sum^{K}_{q=1} d^q_{intra},\\
&&  \mathcal{L}_{inter} = \sum^{K}_{p=1}\sum^{K}_{q=1} \mathbbm{1}[q\not=p]exp(-d^{p,q}_{inter}),
\end{eqnarray}
where $\mathcal{L}_{intra}$ aims at diminishing intra-class distinction under the supervision of GT, and $\mathcal{L}_{inter}$ focuses on enlarging the differences between class centers that can operate without GT. 

The weighted summation of the $\mathcal{L}_{intra}$ and $\mathcal{L}_{inter}$ is called Class Distance~(CD) Loss for convenience, which can be formulated as:
\begin{equation}
    \mathcal{L}_{CD} = \mathcal{L}_{intra} + \alpha \mathcal{L}_{inter},
\end{equation} where $\alpha$ is the a hyper-parameter that set empirically.

For dataset-level CD Loss, we regard each $\mathbf{w}^\top_{j}$ as dataset-level class center just as our rethinking of traditional segmentation networks. So, the dataset-level intra-class distance loss and inter-class distance loss are calculated respectively as:
\begin{eqnarray}
&&   \mathcal{L}^{dataset}_{intra}=\sum^{K}_{q=1} \sum^{N}_{i=1}\mathbbm{1}[y_{iq}=1] D(\mathbf{f}_i, \mathbf{w}^\top_q)\label{intradataset},\ \ \ \ \ \ \ \\
&&   \mathcal{L}^{dataset}_{inter}=\sum^{K}_{p=1} \sum^{K}_{q=1}\mathbbm{1}[q\not=p]exp\big(-D(\mathbf{w}^\top_p, \mathbf{w}^\top_q)\big).\ \ \ \ \ \ \ 
\end{eqnarray}
Please notice that when combining Eq.~\ref{dlog} with Eq.~\ref{dintra}, Eq.~\ref{rsimi} and Eq.~\ref{sip}, the $\mathcal{L}_{intra}$ is in the same format with CE Loss shown in Eq.~\ref{CE}:
\begin{equation}
    \mathcal{L}^{dataset}_{intra} =\sum^{K}_{q=1} \sum^{N}_{i=1}\mathbbm{1}[y_{iq}=1]\Big \{ \mathit{log}\big(\sum_{j }\mathit{exp}(\mathbf{f}_i\cdot\mathbf{c}^\top_j)\big)-\mathbf{f}_i\cdot\mathbf{c}^\top_q\Big \}.
\end{equation}
This shows that the CE Loss is a special case of our proposed intra-class distance loss.

For scene-level CD Loss, we use the proposed ACCM to calculate the class centers conditioned on each scene.
So, the scene-level CD Loss can be formulated as follows:
\begin{eqnarray}
&&   \mathcal{L}^{scene}_{intra} = \sum^{K}_{q=1} \sum^{N}_{i=1}\mathbbm{1}[y_{iq}=1] D(\mathbf{f}_{i}, \mathbf{c}^\top_{q}),~ \ \ \ \ \\
&&   \mathcal{L}^{scene}_{inter} = \sum^{K}_{p=1} \sum^{K}_{q=1}\mathbbm{1}[q\not=p]exp\big(-D(\mathbf{c}^\top_{p}, \mathbf{c}^\top_{q})\big).\ \ \ \
\end{eqnarray}

\paragraph{Overall loss function.}
The over-all loss function of our model are as follows:
\begin{eqnarray}
    \mathcal{L} &=& \mathcal{L}_{CD} + \beta \mathcal{L}_{Dice} \nonumber \\
    &=&\mathcal{L}_{intra} + \alpha \mathcal{L}_{inter} + \beta \mathcal{L}_{Dice},
\end{eqnarray}where $\mathcal{L}_{Dice}$ is the Dice loss imposed on the mask $\mathbf{M}$ and $\beta$ is the weight of $\mathcal{L}_{Dice}$.
Since the scene-level CD Loss is much more powerful, we employ the $\mathcal{L}^{scene}_{intra}$ and $\mathcal{L}^{scene}_{inter}$ instead of $\mathcal{L}^{dataset}_{intra}$ and $\mathcal{L}^{dataset}_{inter}$.
We conduct experiments to show the effectiveness of $\mathcal{L}^{scene}_{intra}$ and $\mathcal{L}^{scene}_{inter}$ over $\mathcal{L}^{dataset}_{intra}$ and $\mathcal{L}^{dataset}_{inter}$ in Tab.~\ref{tab:step}.

\section{Experiment}
\label{sec:exp}
We evaluate our approach on two challenging semantic segmentation datasets: ADE20K and Pascal Context. We perform a comprehensive ablation study on the ADE20K dataset and report the comparison with other methods on the ADE20K validation set and the Pascal Context validation set.

\subsection{Settings}

\paragraph{Dataset.} ADE20K~\cite{cascadenet} dataset is a large-scale scene parsing benchmark with 150 fine-grained objects and stuff categories, containing 20,210 images for training, 2,000 images for validation, and 3352 images for testing.

Pascal Context dataset~\cite{pascalcontext} is a scene parsing dataset that provides semantic labels for whole scene(both ``things" and ``stuff" classes), which augments 10,103 images from PASCAL VOC 2010~\cite{voc2010}. It has 4,998 training and 5,105 validation images. We use the 59 most common categories for evaluation.

\paragraph{Training.} We conduct our experiments using four NVIDIA GTX 2080 ti GPUs with four images per GPU. All of our models are optimized by SGD optimizer with 0.9 momentum. The initial learning rate is set $1e^{-2}$ for ADE20K and $4e^{-3}$ for PASCAL Context. We adopt the polynomial learning rate decay strategy in training following previous works~\cite{LDF,cpnet,deeplabv3+}. The initial learning rate is multiplied by $(1-\frac{iter}{max iter})^{0.9}$. We apply random resizing with a ratio between 0.5 and 2 in training, random cropping input images into (512,512), and random horizontal flipping during training for all the experiments. We set the total iterations on ADE20K to 80K and 160K for our ResNet-50 and ResNet-101 models, respectively. For Pascal Context, models run 80k iterations during training.

\paragraph{Auxiliary loss.} Follow previous work~\cite{crfrnn}, we adopt auxiliary segmentation loss to help train our model. We add an auxiliary FCN head, which outputs prediction under the supervision of CE Loss multiplied by 0.4.

\paragraph{Evaluation.}
During the evaluation, we average the predictions of multiple scaled following the previous work~\cite{psp,cpnet,deeplabv3+}. Each image is then flipped horizontally, then scaled to a uniform size with scaling factor (0.5, 0.75, 1.0, 1.25, 1.5, 1.75) for better performance. Besides, we use Synchronized BN in our models. Additionally, we report mean Intersection over Union (mIoU) and pixel accuracy (Acc) for ADE20K and mIoU for PASCAL context.

\subsection{Ablation Study On ADE20K}
We conduct ablation studies on ADE20K to demonstrate the effectiveness of our approach. Models are trained on ADE20K train set and evaluated on val set.
All the models are pretrained on Image-Net without extra data.

\paragraph{Upper-bound verification.} 
To verify the feasibility of our proposed methods, we first carry out a simple upper-bound verification experiment.
We directly replace the predicted $Mask$, which is supervised by CE loss, of our trained model with processed ground truth during inference. 
According to the results shown in Tab.~\ref{tab:upper}, our method provides a huge improvement compared with the baseline, while the upper bound method Ours-GT greatly outperforms our method
when the predicted $Mask$ is replaced by GT.
Since the GT provides better weight maps than the predicted $Mask$ to generate the adaptive class centers, the performance is dramatically improved, indicating that there is still a lot of room for improvement.

\begin{table}[htpb]
\centering
\small
\tablestyle{7pt}{1.2}
\begin{center}
\begin{tabular}{l|c|c|c}
\shline
Method  &CCS  & mIoU($\%$)& Acc($\%$)\\
\shline
Baseline     &        & 37.94     & 77.98 \\
Ours         & \checkmark       & 43.57     & 81.06 \\
Ours-GT      & \checkmark       & 47.21     & 84.12 \\

\shline
\end{tabular}
\end{center}
\caption{Upper-bound verification. Baseline: ResNet-50 FCN. 
Ours: Baseline + CCS layer.
Ours-GT: Baseline + CCS layer whose $Mask$ is supplanted by GT to verify the upper bound of CCS layer.}
\label{tab:upper}
\end{table}

\begin{figure}[ht]
\begin{center}
\includegraphics[width=0.48\linewidth]{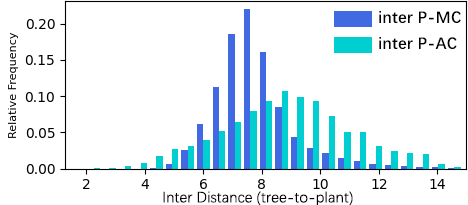}
\includegraphics[width=0.48\linewidth]{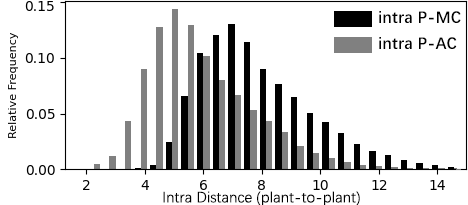}
\end{center}
\caption{P-C \textbf{distance} comparison between MC \& AC.
Left: inter-class P-C distance.
Right: intra-class P-C distance.
We denote pixel-to-mean centers and pixel-to-adaptive centers as MC and AC, respectively.}
\vspace{-0.3cm}
\label{fig:hist}
\end{figure}

\paragraph{Ablation study of class centers.} 
To demonstrate the effectiveness of the adaptive class centers, we compare the P-C similarity, depending on which we generated the probability maps. 
Taking ``tree" and ``plant" as examples, we show the frequency histograms of intra-class pixel-to-adaptive centers distance and pixel-to-mean centers~(namely the global centers that learned from the whole dataset) distance in Fig.~\ref{fig:hist}.
As we introduced before, the results demonstrate that the adaptive class centers make prominent success in increasing the inter-class distance between ``tree" pixels and ``plant" class center, while the intra-class distances of ``plant" pixels and adaptive class center of ``plant" are dramatically smaller than the distance between pixels and global class center.

\paragraph{Ablation study of ACCM.} 
We conduct ablation studies of ACCM to explore the best performance. 
We compared different loss functions on $Mask$ in order to generate better adaptive class centers. We conduct experiments on FCN based on ResNet-50 and report the results in Tab.~\ref{tab:loss}.

\begin{table}[htpb]
\centering
\small
\tablestyle{7pt}{1.2}
\begin{center}
\begin{tabular}{c|c|c|c}
\shline
Res50 FCN CCSNet & w/o loss & w/ CE loss & w/ Dice loss \\
\shline
mIoU($\%$)     & 43.01    & 41.56      & 43.57\\
Acc($\%$)      & 80.69    & 80.45      & 81.06\\
\shline
\end{tabular}
\end{center}
\caption{Ablation study of defferent loss fucntion on $\mathbf{Mask}$. Our baseline in this experiment is ResNet-50 FCN, with our CS replace the final convolution of FCN under supervision of scene-level CD Loss.}
\label{tab:loss}
\vspace{-5pt}
\end{table}

\begin{table}
\centering
\small
\tablestyle{7pt}{1.2}
\begin{center}
\begin{tabular}{l|c|c|c}
\shline
Methods & FCN & PSP & Deeplabv3+ \\
\shline
baseline & 37.94 & 41.94 & 43.57 \\
+CCS & 43.57 & 44.07 & 44.25\\
\shline
\end{tabular}
\end{center}
\caption{Ablation study of segmentation heads. CCS layer is equipped on different segmentation heads based on ResNet-50 are trained on ADE20K training set for 80k iterations. The mIoU results on ADE20K validati set of each model are reported.}
\label{tab:decoder}
\vspace{-5pt}
\end{table}

\paragraph{Ablation study of hyper-parameter and architecture.} We first conduct experiments with different segmentation heads. The multi-scale mIoU results of different segmentation heads based on ResNet-50 are reported in Tab.~\ref{tab:decoder}. Among three different segmentation heads, deeplabv3+ has the best performance with $44.25\%$ ms mIoU. We also explore proper hyper-parameters for CCSNet. We take FCN based on ResNet-50 as a baseline to exploit the best weight of $d_{inter}$ and the weight of Dice loss. As shown in Tab.~\ref{tab:interweight}, our approach achieves the best performance when the weight of $d_{inter}$ is empirically set as 0.5. The best FCN based model equipped with our approach improves the pix-Acc and mIoU by 2.59\% and 5.63\%.

\begin{table}
\centering
\small
\tablestyle{7pt}{1.2}
\begin{center}
\begin{tabular}{c|c|c|c}
\shline
mIoU & $\beta=0.1$ & $\beta=0.5$ & $\beta=1.0$ \\
\shline
$\alpha=0.1$ & 42.79 & 42.86 & 42.85\\
$\alpha=0.5$ & 43.01 & 43.05 & $\mathbf{43.57}$\\
$\alpha=1.0$ & 43.14 & 43.50 & 43.47\\
\shline
\end{tabular}
\end{center}
\caption{Alblation Study of weight of $\mathcal{L}^{scene}_{inter}$ and weight of dice loss. We vary the value of $\alpha$ and $\beta$ and find when $\alpha=0.5$, $\beta=1.0$, model has the best performance.}
\label{tab:interweight}
\vspace{-5pt}
\end{table}

\paragraph{Ablation study of CCS layer.} We break down the improvements of our work over ResNet-101 based on DeeplabV3+, which has the best performance. We add the proposed components in our approaches step by step to the DeeplabV3+ baseline. Experiments are conducted on ADE20K and run 160k iterations. By simply replacing the global class centers $\mathbf{W}^c$ with our adaptive class centers conditionally generated by 
ACCM, our network improves the performance by 0.86\% in mIoU. This provides strong support for our assertion that global class centers are inferior compared to conditional class centers. Moreover, together with the image-level inter-class distance loss, our network achieves 47.76\% mIoU on the ADE20K validation set, which demonstrates the effectiveness of our inter-class distance loss.

\begin{figure*}[htb]
\footnotesize
\centering
\renewcommand{\tabcolsep}{1pt} %
\renewcommand{\arraystretch}{1} %
\begin{center}
\begin{tabular}{ccccc}
\includegraphics[width=0.18\linewidth]{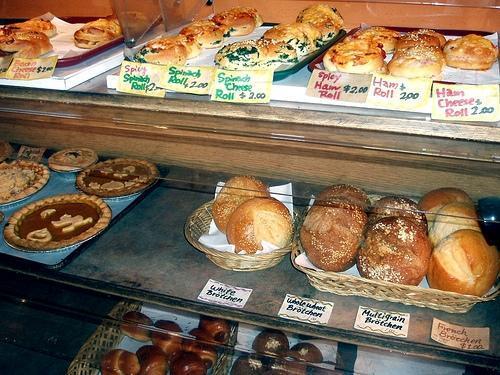} &
\includegraphics[width=0.18\linewidth]{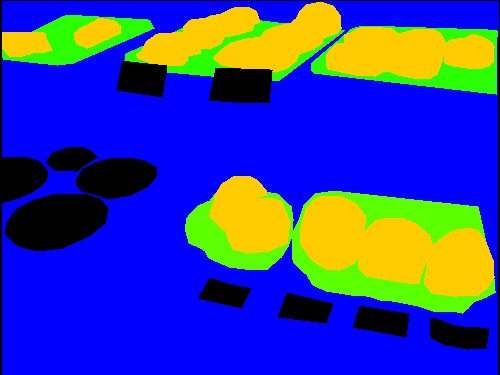} &
\includegraphics[width=0.18\linewidth]{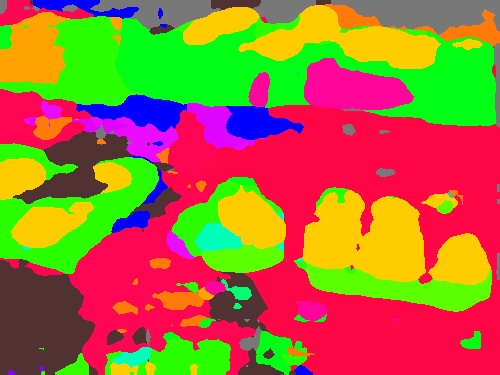}  &
\includegraphics[width=0.18\linewidth]{figs/img2/deep/ADE_val_00000059.jpg}  &
\includegraphics[width=0.18\linewidth]{figs/img2/ours/ADE_val_00000059.jpg}  \\
\includegraphics[width=0.18\linewidth]{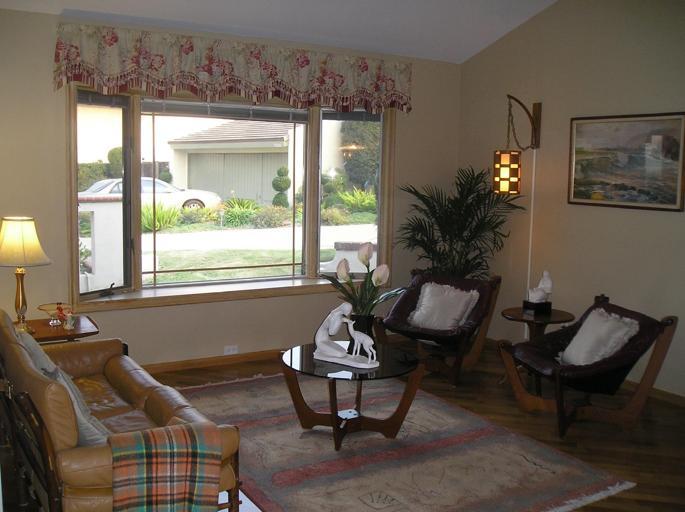}  &
\includegraphics[width=0.18\linewidth]{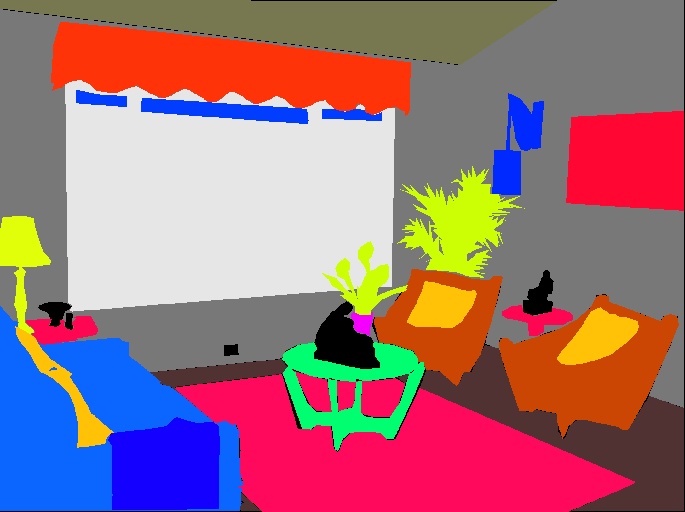}  &
\includegraphics[width=0.18\linewidth]{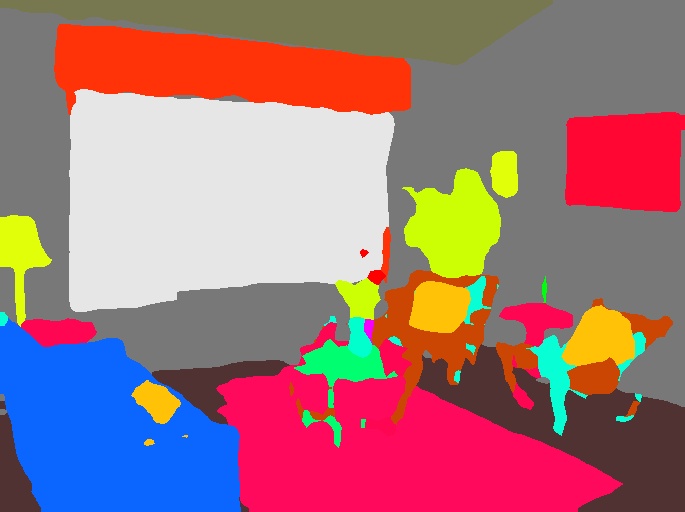}   &
\includegraphics[width=0.18\linewidth]{figs/img2/deep/ADE_val_00001971.jpg}  &
\includegraphics[width=0.18\linewidth]{figs/img2/ours/ADE_val_00001971.jpg}  \\
\includegraphics[width=0.18\linewidth]{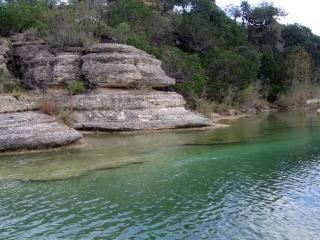}  &
\includegraphics[width=0.18\linewidth]{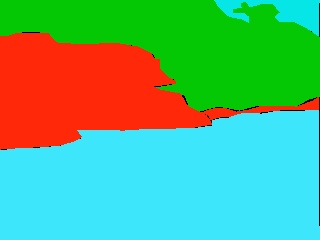}  &
\includegraphics[width=0.18\linewidth]{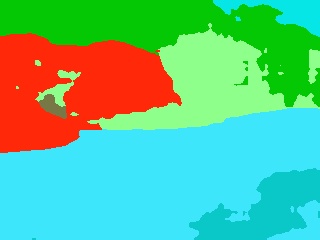}  &
\includegraphics[width=0.18\linewidth]{figs/img2/deep/ADE_val_00001996.jpg}  &
\includegraphics[width=0.18\linewidth]{figs/img2/ours/ADE_val_00001996.jpg}  \\
\includegraphics[width=0.18\linewidth]{figs/img2/img/ADE_val_00000314.jpg}   &
\includegraphics[width=0.18\linewidth]{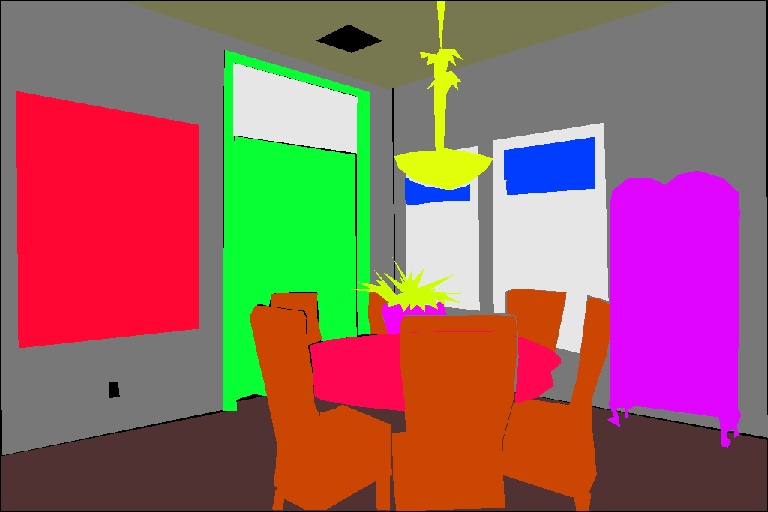}  &
\includegraphics[width=0.18\linewidth]{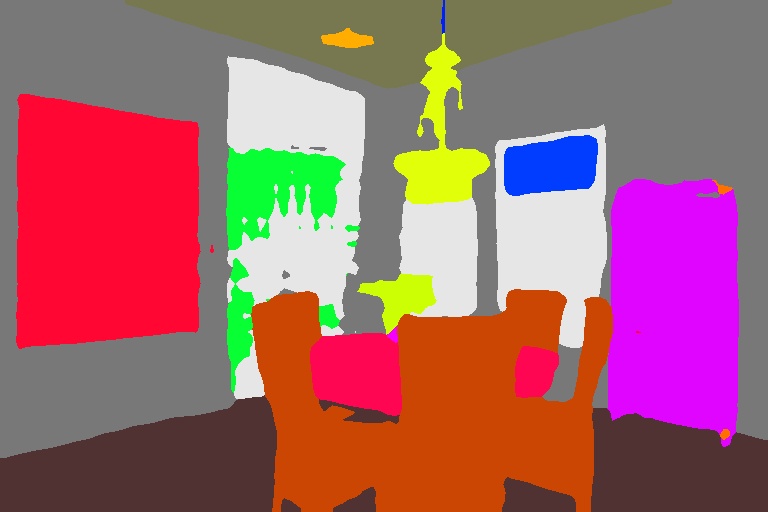}   &
\includegraphics[width=0.18\linewidth]{figs/img2/deep/ADE_val_00000314.jpg}   &
\includegraphics[width=0.18\linewidth]{figs/img2/ours/ADE_val_00000314.jpg}  \\
(a) Images &  (b) GT  & (c) FCN  & (d) DeeplabV3+ & (e) CCSNet~(Ours)\\
\end{tabular}
\end{center}
\caption{
Qualitative results on ADE20K validation set.
}
\vspace{-5pt}
\label{fig:vade}
\end{figure*}

\subsection{Results on ADE20K}
We train the ResNet-101 on the ADE20K training set and report the mIoU results on the validation set (results are shown in Tab.~\ref{tab:sotaade}). Our CCSNet uses a pre-trained backbone network on ImageNet and achieves 47.76\% mIoU, which outperforms DeeplabV3+ by 1.41\% mIoU using the same backbone network. Visual comparison examples are shown in Fig.~\ref{fig:vade}.

\begin{table}
\centering
\small
\tablestyle{7pt}{1.2}
\begin{center}
\begin{tabular}{l|c|c|c|c}
\shline

Method                               & centers & $\mathcal{L}_{inter}$           & $\mathcal{L}_{intra}$ & mIoU \\\shline
Deeplabv3+                               & $\mathbf{W}^c$  & ~                    & $\mathbf{W}^c$ & 46.35\\ \hline
+ $\mathcal{L}^{dataset}_{inter}$        & $\mathbf{W}^c$  & $\mathbf{W}^c$     & $\mathbf{W}^c$ & 46.45\\ 
+ACCM\&$\mathcal{L}^{scene}_{inter}$      & $\mathbf{W}^c$  & $\mathbf{C}$    & $\mathbf{W}^c$ & 46.94\\
+CCS w/o $\mathcal{L}_{inter}$           & $\mathbf{C}$  &                      & $\mathbf{C}$ & 47.21\\
CCSNet~(Ours)                                     & $\mathbf{C}$  & $\mathbf{C}$     & $\mathbf{C}$ & 47.76\\

\shline
\end{tabular}
\end{center}
\caption{Ablation study of CCS layer and CD Loss. The ``centers" are those used to calculate the P-C similarity for prediction. For example, $\mathbf{W}^c$ denotes the learned weights of classifier, namely global class centers, used for final prediction, while the $\mathbf{C}$ indicate that the prediction is based on the P-C similarity using adaptive class centers. ``$\mathcal{L}_{inter}$" and ``$\mathcal{L}_{inter}$" shows whether the loss is applied on $\mathbf{W}^c$~($\mathcal{L}^{dataset}_{inter}$ and $\mathcal{L}^{dataset}_{intra}$) or $\mathbf{C}$~($\mathcal{L}^{scene}_{inter}$ and $\mathcal{L}^{scene}_{intra}$).}
\label{tab:step}
\end{table}

\begin{table}
\centering
\small
\tablestyle{18pt}{1.1}
\begin{center}
\begin{tabular}{l|l|c}
\shline
Method                     & Baseline   & mIoU(\%) \\
\shline
CascadeNet~\cite{cascadenet} & VGG-16 & 34.90 \\
RefineNet~\cite{refinenet}  & ResNet-152 & 40.7 \\
UperNet~\cite{refinenet}  & ResNet-101 & 42.66 \\
PSPNet~\cite{psp}           & ResNet-101 & 43.51    \\
PSPNet~\cite{psp}           & ResNet-269 & 44.94    \\
PSANet~\cite{psa}          & ResNet-269 & 43.77  \\
EncNet~\cite{enc}          & ResNet-101 & 43.77  \\
CFNet~\cite{cfnet}          & ResNet-101 & 44.65    \\
ANL~\cite{ANL}              & ResNet-101 & 45.24    \\
OCRNet~\cite{ocr}          & ResNet-101 & 45.28    \\
APCNet~\cite{apc}           & ResNet-101 & 45.38    \\
RGNet~\cite{rpnet}          & ResNet-101 & 45.80  \\
CPNet~\cite{cpnet}          & ResNet-101 & 46.27  \\
DeeplabV3+~\cite{deeplabv3+}  &ResNet-101 & 46.35 \\ \hline
CCSNet                      &ResNet-101 & $\mathbf{47.76}$ \\
\shline
\end{tabular}
\end{center}
\caption{Results on the ADE20K validation set. Our model based on ResNet-101 achieves 47.76\% in mIoU and outperforms all previous methods using the same backbone network.}
\label{tab:sotaade}
\end{table}

\subsection{Results on PASCAL Context}
Tab.~\ref{tab:sotacontext} reports the comparison results of our network and other state-of-the-the-art approaches. Based on ResNet-101, our method makes favourable performance and achieves 54.9\% mIoU. CCSNet outperforms previous methods using the same ResNet-101 backbone.

\begin{table}
\centering
\small
\tablestyle{18pt}{1.1}
\begin{center}
\begin{tabular}{l|l|c}
\shline
Method                     & Baseline   & mIoU(\%) \\
\shline

RefineNet~\cite{refinenet}  & ResNet-152 & 47.3 \\
PSPNet~\cite{psp}           & ResNet-101 & 47.8    \\
DeeplabV3+~\cite{deeplabv3+}  &ResNet-101 & 48.3 \\
EncNet~\cite{enc}          & ResNet-101 & 51.7  \\
DANet\cite{danet}          & ResNet-101 & 52.6    \\
ANL~\cite{ANL}              & ResNet-101 & 52.8    \\
CFNet~\cite{cfnet}          & ResNet-101 & 54.0    \\
APCNet~\cite{apc}           & ResNet-101 & 54.7    \\
RGNet~\cite{rpnet}          & ResNet-101 & 53.9  \\
CPNet~\cite{cpnet}          & ResNet-101 & 53.9  \\
OCRNet~\cite{ocr}           & ResNet-101 & 54.8    \\
 \hline
CCSNet                      &ResNet-101 & $\mathbf{54.9}$ \\
\shline
\end{tabular}
\end{center}
\caption{Results on the PASCAL Context validation set. We report our result evaluated on 59 class without background. Our model based on ResNet-101 achieves 54.9\% in mIoU.}
\label{tab:sotacontext}
\end{table}

\section{Conclusion}
In this paper, we 
provide a novel perspective to view typical semantic segmentation models, and re-model the problem as a task that models compute the similarity maps between pixels and class centers on each scene, then assign pixels to the semantic category with the highest P-C similarity. Based on this perspective, we provide solutions to address the two typical issues, \ie same category yet different scenes features could be of large variance, while features of different categories but the same scene may be quite similar to each other.
Based on P-C similarity maps, we propose ACCM to generate adaptive class centers conditioned on each scene to deal with the feature variances of different scenes and design inter-class and intra-class distance loss at scene-level for more inter-class distinction inside each scene. Our approach is easy yet effective and can be plugged into most FCN-based architectures. Finally, CCSNet achieves state-of-the-the-art performance on two challenging semantic segmentation datasets.

\bibliographystyle{cvm}

\begin{thebibliography}{10}\itemsep=-1pt

\bibitem{segnet}
V.~Badrinarayanan, A.~Kendall, and R.~Cipolla.
\newblock Segnet: A deep convolutional encoder-decoder architecture for image
  segmentation.
\newblock {\em IEEE transactions on pattern analysis and machine intelligence},
  39(12):2481--2495, 2017.

\bibitem{crf2}
S.~Chandra, N.~Usunier, and I.~Kokkinos.
\newblock Dense and low-rank gaussian crfs using deep embeddings.
\newblock In {\em Proceedings of the IEEE International Conference on Computer
  Vision}, pages 5103--5112, 2017.

\bibitem{deeplab}
L.-C. Chen, G.~Papandreou, I.~Kokkinos, K.~Murphy, and A.~L. Yuille.
\newblock Deeplab: Semantic image segmentation with deep convolutional nets,
  atrous convolution, and fully connected crfs.
\newblock {\em IEEE transactions on pattern analysis and machine intelligence},
  40(4):834--848, 2017.

\bibitem{ASPP}
L.-C. Chen, G.~Papandreou, F.~Schroff, and H.~Adam.
\newblock Rethinking atrous convolution for semantic image segmentation.
\newblock {\em arXiv preprint arXiv:1706.05587}, 2017.

\bibitem{deeplabv3+}
L.-C. Chen, Y.~Zhu, G.~Papandreou, F.~Schroff, and H.~Adam.
\newblock Encoder-decoder with atrous separable convolution for semantic image
  segmentation.
\newblock In {\em Proceedings of the European conference on computer vision
  (ECCV)}, pages 801--818, 2018.

\bibitem{deformable}
J.~Dai, H.~Qi, Y.~Xiong, Y.~Li, G.~Zhang, H.~Hu, and Y.~Wei.
\newblock Deformable convolutional networks.
\newblock In {\em Proceedings of the IEEE international conference on computer
  vision}, pages 764--773, 2017.

\bibitem{voc2010}
M.~Everingham and J.~Winn.
\newblock The pascal visual object classes challenge 2012 (voc2012) development
  kit.
\newblock {\em Pattern Analysis, Statistical Modelling and Computational
  Learning, Tech. Rep}, 8, 2011.

\bibitem{DUALA1}
J.~Fu, J.~Liu, J.~Jiang, Y.~Li, Y.~Bao, and H.~Lu.
\newblock Scene segmentation with dual relation-aware attention network.
\newblock {\em IEEE Transactions on Neural Networks and Learning Systems},
  PP:1--14, 08 2020.

\bibitem{danet}
J.~Fu, J.~Liu, H.~Tian, Y.~Li, Y.~Bao, Z.~Fang, and H.~Lu.
\newblock Dual attention network for scene segmentation.
\newblock In {\em Proceedings of the IEEE/CVF Conference on Computer Vision and
  Pattern Recognition}, pages 3146--3154, 2019.

\bibitem{apc}
J.~He, Z.~Deng, L.~Zhou, Y.~Wang, and Y.~Qiao.
\newblock Adaptive pyramid context network for semantic segmentation.
\newblock In {\em Proceedings of the IEEE/CVF Conference on Computer Vision and
  Pattern Recognition}, pages 7519--7528, 2019.

\bibitem{resnet}
K.~He, X.~Zhang, S.~Ren, and J.~Sun.
\newblock Deep residual learning for image recognition.
\newblock In {\em Proceedings of the IEEE conference on computer vision and
  pattern recognition}, pages 770--778, 2016.

\bibitem{huang2017densely}
G.~Huang, Z.~Liu, L.~Van Der~Maaten, and K.~Q. Weinberger.
\newblock Densely connected convolutional networks.
\newblock In {\em Proceedings of the IEEE conference on computer vision and
  pattern recognition}, pages 4700--4708, 2017.

\bibitem{huang2019ccnet}
Z.~Huang, X.~Wang, L.~Huang, C.~Huang, Y.~Wei, and W.~Liu.
\newblock Ccnet: Criss-cross attention for semantic segmentation.
\newblock In {\em Proceedings of the IEEE/CVF International Conference on
  Computer Vision}, pages 603--612, 2019.

\bibitem{krizhevsky2012imagenet}
A.~Krizhevsky, I.~Sutskever, and G.~E. Hinton.
\newblock Imagenet classification with deep convolutional neural networks.
\newblock {\em Advances in neural information processing systems},
  25:1097--1105, 2012.

\bibitem{li2019global}
X.~Li, L.~Zhang, A.~You, M.~Yang, K.~Yang, and Y.~Tong.
\newblock Global aggregation then local distribution in fully convolutional
  networks.
\newblock {\em arXiv preprint arXiv:1909.07229}, 2019.

\bibitem{refinenet}
G.~Lin, A.~Milan, C.~Shen, and I.~Reid.
\newblock Refinenet: Multi-path refinement networks for high-resolution
  semantic segmentation.
\newblock In {\em Proceedings of the IEEE conference on computer vision and
  pattern recognition}, pages 1925--1934, 2017.

\bibitem{mrf}
Z.~Liu, X.~Li, P.~Luo, C.-C. Loy, and X.~Tang.
\newblock Semantic image segmentation via deep parsing network.
\newblock In {\em Proceedings of the IEEE international conference on computer
  vision}, pages 1377--1385, 2015.

\bibitem{long2015fully}
J.~Long, E.~Shelhamer, and T.~Darrell.
\newblock Fully convolutional networks for semantic segmentation.
\newblock In {\em Proceedings of the IEEE conference on computer vision and
  pattern recognition}, pages 3431--3440, 2015.

\bibitem{dice}
F.~Milletari, N.~Navab, and S.-A. Ahmadi.
\newblock V-net: Fully convolutional neural networks for volumetric medical
  image segmentation.
\newblock In {\em 2016 fourth international conference on 3D vision (3DV)},
  pages 565--571. IEEE, 2016.

\bibitem{pascalcontext}
R.~Mottaghi, X.~Chen, X.~Liu, N.-G. Cho, S.-W. Lee, S.~Fidler, R.~Urtasun, and
  A.~Yuille.
\newblock The role of context for object detection and semantic segmentation in
  the wild.
\newblock In {\em Proceedings of the IEEE Conference on Computer Vision and
  Pattern Recognition}, pages 891--898, 2014.

\bibitem{unet}
O.~Ronneberger, P.~Fischer, and T.~Brox.
\newblock U-net: Convolutional networks for biomedical image segmentation.
\newblock In {\em International Conference on Medical image computing and
  computer-assisted intervention}, pages 234--241. Springer, 2015.

\bibitem{russakovsky2015imagenet}
O.~Russakovsky, J.~Deng, H.~Su, J.~Krause, S.~Satheesh, S.~Ma, Z.~Huang,
  A.~Karpathy, A.~Khosla, M.~Bernstein, et~al.
\newblock Imagenet large scale visual recognition challenge.
\newblock {\em International journal of computer vision}, 115(3):211--252,
  2015.

\bibitem{simonyan2014very}
K.~Simonyan and A.~Zisserman.
\newblock Very deep convolutional networks for large-scale image recognition.
\newblock {\em arXiv preprint arXiv:1409.1556}, 2014.

\bibitem{tsne}
L.~van~der Maaten and G.~Hinton.
\newblock Viualizing data using t-sne.
\newblock {\em Journal of Machine Learning Research}, 9:2579--2605, 11 2008.

\bibitem{attention}
A.~Vaswani, N.~Shazeer, N.~Parmar, J.~Uszkoreit, L.~Jones, A.~N. Gomez,
  L.~Kaiser, and I.~Polosukhin.
\newblock Attention is all you need.
\newblock {\em arXiv preprint arXiv:1706.03762}, 2017.

\bibitem{nonlocal}
X.~Wang, R.~Girshick, A.~Gupta, and K.~He.
\newblock Non-local neural networks.
\newblock In {\em Proceedings of the IEEE conference on computer vision and
  pattern recognition}, pages 7794--7803, 2018.

\bibitem{denseaspp}
M.~Yang, K.~Yu, C.~Zhang, Z.~Li, and K.~Yang.
\newblock Denseaspp for semantic segmentation in street scenes.
\newblock In {\em Proceedings of the IEEE conference on computer vision and
  pattern recognition}, pages 3684--3692, 2018.

\bibitem{rpnet}
C.~Yu, Y.~Liu, C.~Gao, C.~Shen, and N.~Sang.
\newblock Representative graph neural network.
\newblock In {\em European Conference on Computer Vision}, pages 379--396.
  Springer, 2020.

\bibitem{cpnet}
C.~Yu, J.~Wang, C.~Gao, G.~Yu, C.~Shen, and N.~Sang.
\newblock Context prior for scene segmentation.
\newblock In {\em Proceedings of the IEEE/CVF Conference on Computer Vision and
  Pattern Recognition}, pages 12416--12425, 2020.

\bibitem{LDF}
C.~Yu, J.~Wang, C.~Peng, C.~Gao, G.~Yu, and N.~Sang.
\newblock Learning a discriminative feature network for semantic segmentation.
\newblock In {\em Proceedings of the IEEE conference on computer vision and
  pattern recognition}, pages 1857--1866, 2018.

\bibitem{dilation}
F.~Yu and V.~Koltun.
\newblock Multi-scale context aggregation by dilated convolutions.
\newblock {\em arXiv preprint arXiv:1511.07122}, 2015.

\bibitem{ocr}
Y.~Yuan, X.~Chen, and J.~Wang.
\newblock Object-contextual representations for semantic segmentation.
\newblock {\em arXiv preprint arXiv:1909.11065}, 2019.

\bibitem{ocnet}
Y.~Yuan and J.~Wang.
\newblock Ocnet: Object context network for scene parsing.
\newblock {\em arXiv preprint arXiv:1809.00916}, 2018.

\bibitem{zhang2019acfnet}
F.~Zhang, Y.~Chen, Z.~Li, Z.~Hong, J.~Liu, F.~Ma, J.~Han, and E.~Ding.
\newblock Acfnet: Attentional class feature network for semantic segmentation.
\newblock In {\em Proceedings of the IEEE/CVF International Conference on
  Computer Vision}, pages 6798--6807, 2019.

\bibitem{enc}
H.~Zhang, K.~Dana, J.~Shi, Z.~Zhang, X.~Wang, A.~Tyagi, and A.~Agrawal.
\newblock Context encoding for semantic segmentation.
\newblock In {\em Proceedings of the IEEE conference on Computer Vision and
  Pattern Recognition}, pages 7151--7160, 2018.

\bibitem{cfnet}
H.~Zhang, H.~Zhang, C.~Wang, and J.~Xie.
\newblock Co-occurrent features in semantic segmentation.
\newblock In {\em Proceedings of the IEEE/CVF Conference on Computer Vision and
  Pattern Recognition}, pages 548--557, 2019.

\bibitem{psp}
H.~Zhao, J.~Shi, X.~Qi, X.~Wang, and J.~Jia.
\newblock Pyramid scene parsing network.
\newblock In {\em Proceedings of the IEEE conference on computer vision and
  pattern recognition}, pages 2881--2890, 2017.

\bibitem{psa}
H.~Zhao, Y.~Zhang, S.~Liu, J.~Shi, C.~C. Loy, D.~Lin, and J.~Jia.
\newblock Psanet: Point-wise spatial attention network for scene parsing.
\newblock In {\em Proceedings of the European Conference on Computer Vision
  (ECCV)}, pages 267--283, 2018.

\bibitem{crfrnn}
S.~Zheng, S.~Jayasumana, B.~Romera-Paredes, V.~Vineet, Z.~Su, D.~Du, C.~Huang,
  and P.~H. Torr.
\newblock Conditional random fields as recurrent neural networks.
\newblock In {\em Proceedings of the IEEE international conference on computer
  vision}, pages 1529--1537, 2015.

\bibitem{cascadenet}
B.~Zhou, H.~Zhao, X.~Puig, S.~Fidler, A.~Barriuso, and A.~Torralba.
\newblock Scene parsing through ade20k dataset.
\newblock In {\em Proceedings of the IEEE conference on computer vision and
  pattern recognition}, pages 633--641, 2017.

\bibitem{ANL}
Z.~Zhu, M.~Xu, S.~Bai, T.~Huang, and X.~Bai.
\newblock Asymmetric non-local neural networks for semantic segmentation.
\newblock In {\em Proceedings of the IEEE/CVF International Conference on
  Computer Vision}, pages 593--602, 2019.

\end{thebibliography}

\end{document}